\documentclass[11pt,twocolumn]{article}
\usepackage[utf8]{inputenc}
\usepackage[T1]{fontenc}
\usepackage{lmodern}
\usepackage[margin=1in]{geometry}
\usepackage{mathtools}
\usepackage{amssymb}
\usepackage{graphicx}
\usepackage{array}
\usepackage{multirow}
\usepackage{placeins}
\usepackage{comment}
\usepackage{algorithm}
\usepackage{algpseudocode}
\usepackage{booktabs}
\usepackage{caption}
\usepackage[sort&compress,numbers]{natbib}
\usepackage[hidelinks]{hyperref}

\let\originalincludegraphics\includegraphics
\newsavebox{\arxivfigurebox}
\renewcommand{\includegraphics}[2][]{%
  \sbox{\arxivfigurebox}{\originalincludegraphics[#1]{#2}}%
  \ifdim\wd\arxivfigurebox>\linewidth
    \resizebox{\linewidth}{!}{\usebox{\arxivfigurebox}}%
  \else
    \usebox{\arxivfigurebox}%
  \fi
}

\let\originalappendix\appendix
\renewcommand{\appendix}{%
  \originalappendix
  \numberwithin{equation}{section}
  \renewcommand{\theequation}{\Alph{section}\arabic{equation}}%
}
\newcommand{\keywords}[1]{\par\medskip\noindent\textbf{Keywords:} #1\par\medskip}
\newcommand{\tabletextsize}{\fontsize{7}{8.5}\selectfont}
\DeclareCaptionFont{tabletext}{\tabletextsize}
\newsavebox{\manuscripttablebox}
\newcommand{\fitTableToWidth}[2]{%
    \begingroup
    \tabletextsize
    \sbox{\manuscripttablebox}{#2}%
    \ifdim\wd\manuscripttablebox>#1\relax
        \resizebox{#1}{!}{\usebox{\manuscripttablebox}}%
    \else
        \usebox{\manuscripttablebox}%
    \fi
    \endgroup
}

\hypersetup{
  pdfauthor={Noah J. Bagazinski, Cyril Picard, Faez Ahmed},
  pdftitle={GenPFN: An In-Context Tabular Foundation Model Approach for Augmenting Small Datasets in Engineering},
  pdfkeywords={In-Context Learning, Engineering Data Augmentation, Candidate Design Generation, Tabular Foundation Models, Gradient-Guided Sampling},
  pdfsubject={TabPFN-guided candidate generation, evaluation, and engineering dataset augmentation}
}
\title{GenPFN: An In-Context Tabular Foundation Model Approach for Augmenting Small Datasets in Engineering}
\author{%
  Noah J. Bagazinski\thanks{Corresponding Author.}\quad Cyril Picard\quad Faez Ahmed\\[0.75em]
  \small Department of Mechanical Engineering,\\
  \small Massachusetts Institute of Technology,\\
  \small 77 Massachusetts Ave,\\
  \small Cambridge, MA USA\\
  \small email: \texttt{noahbagz@mit.edu}; \texttt{cyrilp@mit.edu}; \texttt{faez@mit.edu}
}
\date{}
\begin{document}
\maketitle
\begin{abstract}
Costly simulations and experiments limit the data available for engineering design. This paper proposes GenPFN, a training-free candidate generator guided by a pretrained tabular Prior-Data Fitted Network (TabPFN) classifier with fixed weights. The context contains evaluated designs labeled desired (Class~1) or undesired (Class~0). GenPFN interpolates between desired designs and refines candidates using gated classifier gradients and Gaussian noise. Each candidate is evaluated using the problem's constraints and objective functions. Evaluated desired designs and their objective values augment the context for regression training.

Evaluation spans fourteen selected problems, including a two-parameter benchmark and a 44-parameter hull problem. In paired comparisons, GenPFN with TabPFN v1 achieves higher class satisfaction and lower realism distance than the tested adversarial, variational-autoencoder and diffusion generators. The variational-autoencoder generator provides greater coverage and diversity. Relative to the synthetic minority over-sampling technique (SMOTE), GenPFN provides lower coverage distance and greater diversity. Among observed paired regression fits, augmentation reduces estimated AutoGluon validation mean squared error by 62.1\% relative to context-only training. A hull-stern case study yields 99.3\% propeller-compatible candidates versus 1.4\% among randomly sampled context designs. Generation followed by evaluation produces sample sets that augment limited datasets and support training more accurate predictive models.
\end{abstract}

\section{Introduction}
\begin{figure*}[!th]
    \centering
    \includegraphics[width=6.0in]{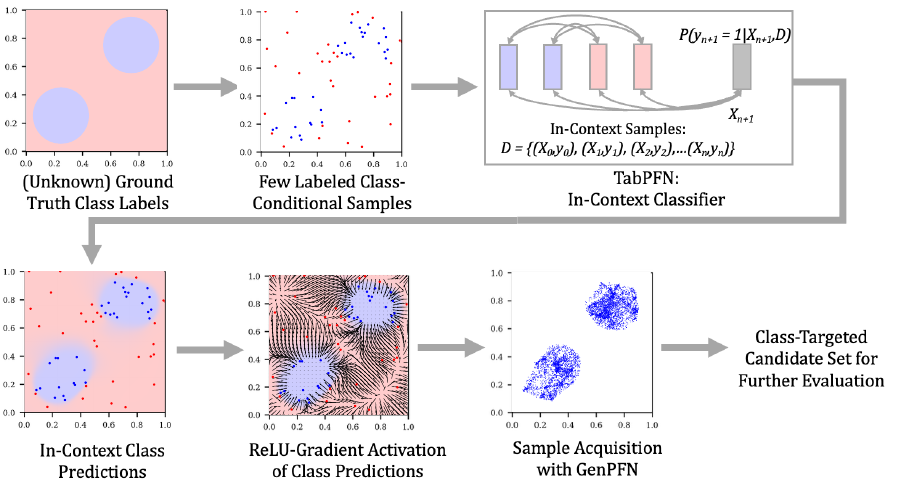}
\caption{GenPFN candidate generation. A pretrained TabPFN v1 classifier uses evaluated designs and their class labels as context while its weights remain fixed. Candidates begin between observed Class~1 designs. Gated classifier gradients direct their updates, gradient clipping limits the directed movement, and Gaussian noise adds variation. Arrows show classifier guidance before clipping and noise are applied. Blue markers represent proposed designs; subsequent problem-specific evaluation determines whether they satisfy the target requirements.}
    \label{fig_1}
\end{figure*}

Engineering design datasets often begin with only 10--1,000 evaluated designs because simulations and experiments are costly~\citep{picard2023dated}. These small datasets must support performance prediction and exploration of high-dimensional, constrained, and multiobjective design spaces. Each additional design consumes evaluation resources, so candidate selection determines which data are added. A method that proposes candidates likely to satisfy the intended engineering requirements can make better use of these resources.

Making those proposals from a small dataset is difficult. Random sampling can spend much of the evaluation budget outside the desired design region, while high-capacity generative models can overfit the few available examples. Candidate generation must also produce varied satisfactory designs to preserve alternatives for subsequent decisions. Set-based design depends on these alternatives as design knowledge matures and can substantially shorten ship-design cycles~\citep{singer2009set,page2022case}. The acquisition challenge is therefore to target a desired class while expanding the range of designs represented in a small dataset.

Tabular foundation models use a set of samples provided as context, similar to a text prompt provided to a large language model. For a Prior-Data Fitted Network (PFN) classifier, this context contains evaluated designs and their class labels. Providing context requires no additional training, and the pretrained PFN weights remain fixed throughout the sampling process~\citep{hollmann2025accurate,qu2025tabicl}. Input gradients indicate how changes to the design variables affect the predicted class probability. Existing PFN generators combine this guidance with noise to generate tabular samples~\citep{ma2023tabpfgen,margeloiu2024tabebm}.

For engineering data augmentation, the goal is to generate varied designs that satisfy the target requirements and obtain their evaluated objective values for predictive-model training.

This paper proposes GenPFN, a training-free candidate generator guided by a pretrained TabPFN classifier whose weights remain fixed. The context labels desired designs as \emph{Class~1} and undesired designs as \emph{Class~0}. GenPFN initializes candidates by interpolating between observed Class~1 designs. During each update, classifier gradients guide candidates whose predicted Class~1 probability falls below a threshold. Gaussian noise adds variation at every update, including when gradient guidance is inactive.

Engineering design exploration can target desired traits even when the corresponding region is difficult to delineate explicitly in the design variables. Labeled examples provide a practical specification of this region of interest. GenPFN uses these examples to acquire additional evaluated designs that support further exploration and predictive modeling within the desired class.

Figure~\ref{fig_1} illustrates candidate generation within the generate--evaluate--augment workflow. Each proposed design is evaluated using the problem's constraints and objective functions to determine its class and objective values. Designs classified as Class~1, together with their evaluated objective values, are added to the original feasible training data for regression.

The study evaluates GenPFN on fourteen problems: a two-parameter benchmark, a 44-parameter hull design problem, and twelve additional engineering problems. The evaluation measures whether generated designs satisfy the target requirements, how their distribution and diversity compare with reference designs, and whether adding evaluated designs improves regression accuracy. A separate hull-stern application examines the generation of propeller-compatible geometries.

The paper makes four contributions:
\begin{enumerate}
    \item A training-free candidate generator combining interpolation, ReLU-activated gradient guidance from a frozen TabPFN classifier, and Gaussian perturbations.
    \item A generate--evaluate--augment workflow that uses externally classified Class~1 designs and their recorded responses to expand regression training data.
    \item A fourteen-problem benchmark using feasibility, realism, coverage, diversity, and downstream regression metrics. A secondary factorial analysis compares the initialization and gradient-gating configurations used within GenPFN.
    \item A ship-stern proof of concept demonstrating class-targeted acquisition of propeller-compatible geometries from a design region that is rarely reached by random sampling.
\end{enumerate}

\section{Background and Prior Work}
GenPFN builds on methods for tabular generation, learning from small labeled datasets, and classifier-guided acquisition of evaluated engineering designs.
\subsection{Generative Artificial Intelligence for Tabular Engineering Design}
Parametric design vectors describe engineering designs using a compact set of variables. Tabular generators produce new values for these variables, and the resulting vectors can drive code-based three-dimensional models~\citep{bagazinski2023shipD,regenwetter2022biked}.

Common tabular generators include the conditional tabular generative adversarial network (CTGAN), tabular variational autoencoder (TVAE), and tabular denoising diffusion probabilistic model (TabDDPM)~\citep{xu2019CTGAN,kotelnikov2023tabddpm}. These models learn a design distribution through task-specific training and have supported applications including airfoils, battery cooling systems, and ship hulls~\citep{chen2020airfoil,chen2021padgan,sung2024cooling,bagazinski2023shipgen,bagazinski2024cshipgen}. They provide the adversarial, variational, and diffusion-based baselines in this study. More recent alternatives include REaLTabFormer, adversarial random forests (ARFs), GOGGLE, and TabSyn~\citep{solatorio2023realtabformer,watson2023arf,liu2023goggle,zhang2024tabsyn}.

For engineering design, these generators should also produce candidates that satisfy the intended constraints or performance criteria. Evaluated datasets often contain examples that both satisfy and violate these requirements. Including negative examples in generative-model training can improve constraint satisfaction~\citep{giannone2023learning,regenwetter2024negative}. Conditional models~\citep{xu2019CTGAN} and classifier-guided tabular diffusion~\citep{bagazinski2023shipgen} also direct generation toward a desired class. Their usefulness depends on how well the available evaluated designs support this targeting.

\subsection{Machine Learning on Small Tabular Data}
Obtaining evaluated engineering designs requires expertise, time, and computational resources~\citep{heyrani2022links,picard2023dated}. These costs often leave few examples available for training a generator. The benchmark therefore compares CTGAN, TVAE and TabDDPM under the small-dataset conditions used by GenPFN.

The same data shortage affects performance prediction. Gradient-boosted tree libraries such as extreme gradient boosting (XGBoost) and categorical boosting (CatBoost) remain strong predictive baselines for small tabular datasets~\citep{chen_xgboost_2016,prokhorenkova2018catboost,SHWARTZZIV202284,regenwetter_deep_2022}. Deep networks often require regularization, while ensembles such as AutoGluon can reduce overfitting and model-selection risk~\citep{srivastava2014dropout,ying2019overview,erickson_autogluon-tabular_2020}. Although these models can learn from few examples, predictions remain difficult in sparsely sampled design regions. Acquiring additional evaluated designs expands the data available for regression training.

\subsection{Data Augmentation and Evaluated Candidate Acquisition}
Data augmentation expands a training set using relationships among existing observations. The synthetic minority over-sampling technique (SMOTE) and adaptive synthetic sampling (ADASYN) address class imbalance through interpolation~\citep{chawla2002smote,he2008adasyn}. SMOTE interpolates between neighboring same-class observations, whereas ADASYN emphasizes difficult-to-classify observations. Mixup interpolates both inputs and labels~\citep{zhang2018mixup}.

Generating a design vector does not supply the objective value needed for regression training. Each candidate must therefore undergo problem-specific evaluation. Class~1 designs and their evaluated objective values then augment the regression training data.

Class-targeted acquisition has supported engineering generative models and Bayesian optimization~\citep{giannone2023learning,yu2024fast,bagazinski2023shipgen,heyrani2021range}. Earlier classifier-guided sampling trained Bayesian network classifiers on evaluated designs and used preferred-class probabilities to direct sequential sampling in discrete, discontinuous engineering spaces~\citep{backlund2012classifier,backlund2015classifier}. GenPFN uses the same principle of directing evaluations through class predictions. Its frozen differentiable PFN predicts class membership from the labeled context. Input gradients then guide updates to batches of candidates.

\subsection{In-context Learning with Tabular Data}
Tabular foundation models learn classification and regression procedures during pretraining. In-context learning applies these procedures to samples supplied for a new problem without additional training or changes to the model weights~\citep{hollmann2025accurate,qu2025tabicl}. Applications include engineering classification, Bayesian inference, constrained Bayesian optimization, and TabPFN v2 density estimation~\citep{picard2024untrained,muller2021transformers,yu2024fast,hollmann2025accurate}. Wang et al.\ also use TabPFN v2 for zero-shot sequential generation and design inpainting~\citep{wang2026tabpfn}; their procedure can depend on variable order.

Differentiable classifier predictions provide another route to generation. Grathwohl et al.\ reinterpreted classifier logits as joint and marginal energy models and sampled them using stochastic gradient Langevin dynamics (SGLD), while Dhariwal and Nichol used target-class log-probability gradients to guide diffusion sampling~\citep{grathwohl2019classifier,dhariwal2021diffusion}. TabPFGen and TabEBM apply classifier-derived energies to frozen-PFN tabular generation~\citep{ma2023tabpfgen,margeloiu2024tabebm}. TabPFGen defines a shared class-conditional energy from TabPFN logits and initializes samples near observed data. TabEBM constructs a separate surrogate binary context and class-specific energy for each target class. Both update complete feature vectors with Langevin-style gradient and noise steps.

GenPFN combines interpolation between observed Class~1 designs with threshold-gated classifier guidance and Gaussian perturbations. Problem-specific evaluation then supplies the class labels and objective values used to augment regression training. The Appendix compares these choices with shared guidance-free controls and local TabPFGen and TabEBM activation variants on TabPFN v1 and v3.

\subsection{SGLD and Langevin-Inspired Sampling}
Langevin-style sampling combines directed movement with random perturbations. The gradient directs samples toward regions favored by a sampling distribution, while noise introduces variation. Equation~\ref{eq_1} gives a representative Bayesian SGLD update:
\begin{equation}\label{eq_1}
\begin{aligned}
\boldsymbol{q}_{n+1}
&=
\boldsymbol{q}_{n}
+\frac{\eta_n}{2}\widehat{\nabla}_{\boldsymbol{q}}\log \pi(\boldsymbol{q}_{n})
+\sqrt{\eta_n}\boldsymbol{\xi}_{n},\\
&
\boldsymbol{\xi}_{n}\sim\mathcal{N}(\boldsymbol{0},\boldsymbol{I}),
\end{aligned}
\end{equation}
where $\pi$ is the target probability density and $\widehat{\nabla}\log\pi$ is commonly a stochastic estimate of the gradient of its log density. The step size $\eta_n$ controls both the directed movement and the Gaussian perturbation~\citep{welling2011bayesian,neal2011mcmc}. Under appropriate regularity conditions and a suitable decreasing step-size schedule, this construction can be used for posterior sampling.

GenPFN uses a similar update structure, with independently chosen guidance and noise scales. Fixed coefficients $\alpha$ and $\sigma$ control the transformed, ReLU-gated guidance and Gaussian noise. The procedure specifies no target density, burn-in, or convergence criterion. It is an \emph{SGLD-inspired} candidate generator and does not perform Markov chain Monte Carlo (MCMC) or posterior sampling.

\section{GenPFN: Class-Conditional Candidate Generation with TabPFN}
GenPFN uses a frozen TabPFN binary classifier to propose engineering designs for a desired class, building on prior PFN-based generators~\citep{ma2023tabpfgen,margeloiu2024tabebm}. It begins by interpolating between observed Class~1 designs, then updates the candidates with classifier gradients and Gaussian noise in normalized design coordinates. The candidates return to physical coordinates for evaluation by the problem-specific constraints and objectives. Constraint-feasible candidates and their externally evaluated responses augment the original feasible context for regression training.

Generation uses normalized coordinates so that design variables with different units and ranges share a common scale. For a physical design vector $\boldsymbol{x}\in\mathbb{R}^{d}$ with parameter limits $\boldsymbol{\ell}$ and $\boldsymbol{u}$, Equation~\ref{eq_2} defines the min--max transformation:
\begin{equation}\label{eq_2}
\boldsymbol{z}
=
(\boldsymbol{x}-\boldsymbol{\ell})\oslash(\boldsymbol{u}-\boldsymbol{\ell}),
\end{equation}
where $\oslash$ denotes elementwise division. The transformed designs and their labels form the context $\mathcal{D}_{\mathrm{ic}}=\{(\boldsymbol{z}_i,y_i)\}_{i=1}^{N}$, with $y_i\in\{0,1\}$ and Class~1 as the target. Conditioned on this context, TabPFN returns the predicted Class~1 probability
$p_{\theta}(\boldsymbol{z}\mid\mathcal{D}_{\mathrm{ic}})=\Pr_{\theta}(y=1\mid\boldsymbol{z},\mathcal{D}_{\mathrm{ic}})$.

At each update, GenPFN predicts the candidate's Class~1 probability. If that probability is below the threshold, classifier guidance supplies a directed step. Gaussian noise is then added regardless of whether guidance is active. The probability and guidance are recomputed at the next update.

Let $\boldsymbol{h}_{R}$ denote the raw gradient obtained with the rectified linear unit (ReLU) gate described below. The update stores the negative of the probability-increasing gradient. Subtracting this stored gradient therefore directs the candidate toward higher predicted Class~1 probability when guidance is active. Equations~\ref{eq_3} and~\ref{eq_4} transform each candidate's raw gradient independently. For a row gradient $\boldsymbol{h}$, per-candidate L2 clipping gives
\begin{equation}\label{eq_3}
\mathcal{C}_{\lambda}(\boldsymbol{h})
=
\frac{\lambda\boldsymbol{h}}
{\max\!\left(\lambda,\lVert\boldsymbol{h}\rVert_{2}\right)}.
\end{equation}
The deterministic step transformation is
\begin{equation}\label{eq_4}
\begin{aligned}
&\mathcal{S}_{\lambda}(\boldsymbol{h})
=\alpha\mathcal{C}_{\lambda}(\boldsymbol{h}),\\
&
\lVert\mathcal{S}_{\lambda}(\boldsymbol{h})\rVert_2
=\alpha\min(\lVert\boldsymbol{h}\rVert_2,\lambda)
\leq\alpha\lambda.
\end{aligned}
\end{equation}
The clipping threshold $\lambda$ limits the gradient norm, and the step coefficient $\alpha$ scales the clipped gradient. Their product, $\alpha\lambda$, bounds the length of the deterministic update. The noise scale $\sigma$ controls the standard deviation of each coordinate's Gaussian perturbation in normalized coordinates. The deterministic bound does not apply to the added noise. Clipping preserves the gradient direction, treats both signs symmetrically, leaves gradients below $\lambda$ unchanged, and maps a zero gradient to zero. The positive coefficients $\alpha$ and $\lambda$ are specified for each problem in the experimental procedure. All PFN sampling configurations use this per-candidate L2 rule.

To begin the trajectory, GenPFN independently selects two Class~1 context vectors, $\boldsymbol{z}_a$ and $\boldsymbol{z}_b$, interpolates between them, and adds an initial Gaussian perturbation. It then performs 100 updates, each combining the transformed gradient with a fresh noise draw. Equations~\ref{eq_5} and~\ref{eq_6} define the initial state and subsequent updates:
\begin{align}
\boldsymbol{z}_{0}
&=
\boldsymbol{z}_{a}+r(\boldsymbol{z}_{b}-\boldsymbol{z}_{a})
+\sigma\boldsymbol{\xi}^{\mathrm{init}},\notag\\
&\qquad r\sim\mathcal{U}(0,1),
\label{eq_5}\\
\boldsymbol{z}_{n+1}
&=
\boldsymbol{z}_{n}
-\mathcal{S}_{\lambda}\!\left[\boldsymbol{h}_{R}(\boldsymbol{z}_{n})\right]
+\sigma\boldsymbol{\xi}_{n},\notag\\
&\qquad n=0,\ldots,99,
\label{eq_6}
\end{align}
where $\boldsymbol{\xi}^{\mathrm{init}}$ and each $\boldsymbol{\xi}_{n}$ are independent draws from $\mathcal{N}(\boldsymbol{0},\boldsymbol{I}_{d})$. Each perturbation has componentwise standard deviation $\sigma$ in normalized coordinates. The benchmark used $\sigma=5\times10^{-4}$ except for Hull, which used $\sigma=1\times10^{-4}$. Thus, every candidate receives one perturbation after interpolation and another at each of the 100 updates. Updates are not projected back into the normalized parameter box. The final iterate supplies the candidate; external evaluation does not select an intermediate state.

Algorithm~\ref{alg_1} summarizes this sequence for a complete candidate set.

\begin{algorithm}[!th]
\caption{GenPFN candidate-generation procedure.}
\begin{algorithmic}[1]
\State \textbf{Input:} Labeled normalized context $\mathcal{D}_{\mathrm{ic}}$, candidate count $M$, $T=100$, $P=0.5$, $\delta=10^{-6}$, step coefficient $\alpha$, clipping threshold $\lambda$, and noise scale $\sigma$
\State \textbf{Output:} Normalized candidate set targeted toward Class~1
\For{$m\gets1$ to $M$}
    \State Select $\boldsymbol{z}_a$ and $\boldsymbol{z}_b$ independently from the Class~1 context vectors
    \State Draw $r\sim\mathcal{U}(0,1)$ and $\boldsymbol{\xi}^{\mathrm{init}}_m\sim\mathcal{N}(\boldsymbol{0},\boldsymbol{I}_d)$
    \State $\boldsymbol{z}^{(0)}_m\gets\boldsymbol{z}_a+r(\boldsymbol{z}_b-\boldsymbol{z}_a)+\sigma\boldsymbol{\xi}^{\mathrm{init}}_m$
    \For{$n\gets0$ to $T-1$}
        \State $p_n\gets p_{\theta}(\boldsymbol{z}^{(n)}_m\mid\mathcal{D}_{\mathrm{ic}})$
        \State $\displaystyle
        \boldsymbol{h}_n =
        \begin{cases}
        -\nabla_{\boldsymbol{z}^{(n)}_m}p_n/(p_n+\delta), & p_n<P,\\
        \boldsymbol{0}, & p_n\geq P
        \end{cases}$
        \State Draw $\boldsymbol{\xi}_{m,n}\sim\mathcal{N}(\boldsymbol{0},\boldsymbol{I}_d)$
        \State $\boldsymbol{z}^{(n+1)}_m=\boldsymbol{z}^{(n)}_m-\mathcal{S}_{\lambda}(\boldsymbol{h}_n)+\sigma\boldsymbol{\xi}_{m,n}$
    \EndFor
\EndFor
\State \textbf{Return:} $\{\boldsymbol{z}^{(T)}_m\}_{m=1}^{M}$
\end{algorithmic}
\label{alg_1}
\end{algorithm}

\subsection{Log-Activated and ReLU-Activated Gradients}

The raw gradient used in the update in Equation~\ref{eq_6} depends on when classifier guidance is active. The log-activated configuration uses guidance at every update. GenPFN's ReLU-activated configuration uses the same field only when the predicted Class~1 probability is below $P=0.5$. Equations~\ref{eq_7}--\ref{eq_11} define these fields and the raw gradients passed to the update. Gaussian noise is added independently in both configurations.

\begin{figure}[h]
\centering\includegraphics[width=0.95\linewidth]{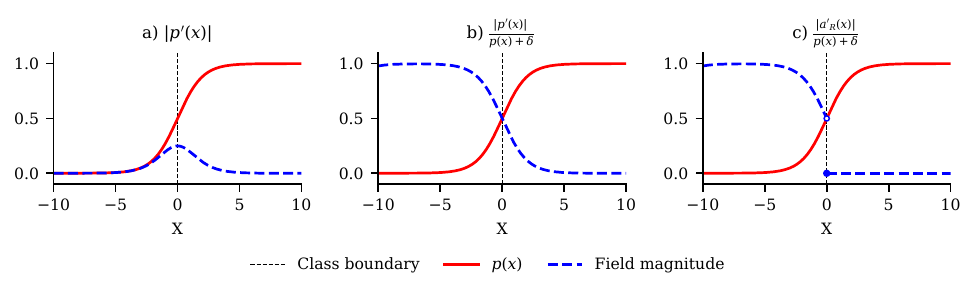}
\caption{One-dimensional gradient-activation rules. The red curve shows the predicted Class~1 probability $p(x)$; the blue dashed curves show (a) $|p'(x)|$, (b) the log-activated magnitude $|p'(x)/(p(x)+\delta)|$, and (c) the ReLU-activated magnitude used by GenPFN. The graphic omits $\delta=10^{-6}$ because it is not visible at this scale, but Equations~\ref{eq_7} and~\ref{eq_10} retain it. The ReLU-activated field equals the log-activated field below $P=0.5$ and is zero otherwise. Both fields precede the transformation in Equation~\ref{eq_4}.}
\label{fig_2}
\end{figure}

Equation~\ref{eq_7} defines the log-activated gradient as the gradient of the stabilized logarithm of the predicted Class~1 probability in normalized coordinates:
\begin{equation}\label{eq_7}
\begin{aligned}
\boldsymbol{g}_{L}(\boldsymbol{z})
&=
\nabla_{\boldsymbol{z}}\log\!\left[
p_{\theta}(\boldsymbol{z}\mid\mathcal{D}_{\mathrm{ic}})+\delta
\right]\\
&=
\frac{\nabla_{\boldsymbol{z}}p_{\theta}(\boldsymbol{z}\mid\mathcal{D}_{\mathrm{ic}})}
{p_{\theta}(\boldsymbol{z}\mid\mathcal{D}_{\mathrm{ic}})+\delta},
\qquad \delta=10^{-6}.
\end{aligned}
\end{equation}
This construction follows the classifier-as-energy and classifier-guidance literature~\citep{grathwohl2019classifier,dhariwal2021diffusion}. The field $\boldsymbol{g}_{L}$ points locally toward increasing stabilized log Class~1 probability. The constant $\delta$ prevents division by zero and limits amplification near zero probability. For a given local probability gradient, the scaling increases the guidance signal as predicted Class~1 probability decreases. Figure labels such as \emph{Log-Gradient} and \emph{Log-Gradient Activation} abbreviate the term \emph{log-activated gradient}.

Equation~\ref{eq_8} defines the stored raw gradient as the negative of this field:
\begin{equation}\label{eq_8}
\boldsymbol{h}_{L}(\boldsymbol{z})
=
-\boldsymbol{g}_{L}(\boldsymbol{z}).
\end{equation}
The log-activated configuration uses $\boldsymbol{h}_{L}$ from Equation~\ref{eq_8} in place of $\boldsymbol{h}_{R}$ in the update in Equation~\ref{eq_6}.

GenPFN gates the same field using the ReLU activation in Equation~\ref{eq_9}, with threshold $P=0.5$:
\begin{equation}\label{eq_9}
a_R(\boldsymbol{z})
=
\operatorname{ReLU}\!\left[P-p_{\theta}(\boldsymbol{z}\mid\mathcal{D}_{\mathrm{ic}})\right].
\end{equation}
Below the threshold, $\nabla_{\boldsymbol{z}}a_R=-\nabla_{\boldsymbol{z}}p_{\theta}$. At or above the threshold, the implementation uses zero gradient. Applying the probability scaling from Equation~\ref{eq_7} gives the \emph{ReLU-activated gradient} in Equation~\ref{eq_10}:
\begin{equation}\label{eq_10}
\begin{aligned}
\boldsymbol{g}_{R}(\boldsymbol{z})
&=
-\frac{\nabla_{\boldsymbol{z}}a_R(\boldsymbol{z})}
{p_{\theta}(\boldsymbol{z}\mid\mathcal{D}_{\mathrm{ic}})+\delta}\\
&=
\begin{cases}
\boldsymbol{g}_{L}(\boldsymbol{z}),
& p_{\theta}(\boldsymbol{z}\mid\mathcal{D}_{\mathrm{ic}})<P,\\[4pt]
\boldsymbol{0},
& p_{\theta}(\boldsymbol{z}\mid\mathcal{D}_{\mathrm{ic}})\geq P.
\end{cases}
\end{aligned}
\end{equation}
Guidance therefore uses the log-activated field below $P$ and switches off at or above $P$. GenPFN evaluates the gate at each update and adds Gaussian noise regardless of the gate's state. Figure labels such as \emph{ReLU-Gradient} and \emph{ReLU-Gradient Activation} abbreviate the term \emph{ReLU-activated gradient}.

\begin{figure*}[!th]
\centering\includegraphics[width=6.0in]{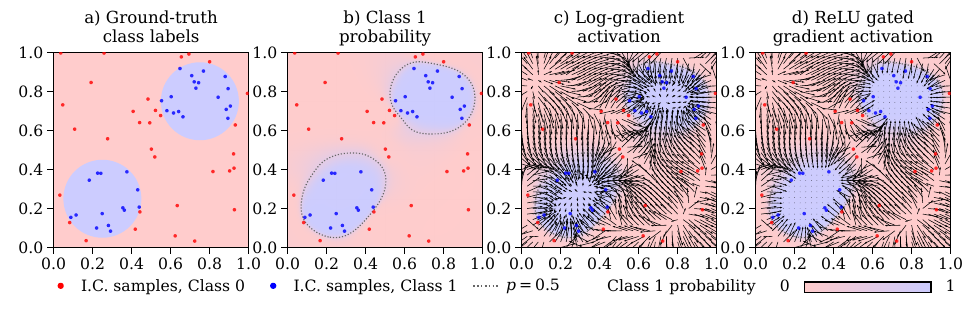}
\caption{TabPFN fields for the \emph{Two Circles} problem: (a) in-context samples and ground-truth classes, (b) predicted Class~1 probability, (c) the log-activated field $\boldsymbol{g}_{L}$, and (d) the ReLU-activated field $\boldsymbol{g}_{R}$, which is zero for $p_{\theta}\geq P=0.5$. The embedded activation labels abbreviate these fields. The fields are computed before $\mathcal{S}_{\lambda}$ and Gaussian noise, so they do not show final candidate displacements.}
\label{fig_3}
\end{figure*}

Equation~\ref{eq_11} defines the raw ReLU-activated gradient using the same sign convention as the log-activated configuration:
\begin{equation}\label{eq_11}
\boldsymbol{h}_{R}(\boldsymbol{z})
=
-\boldsymbol{g}_{R}(\boldsymbol{z})
=
\frac{\nabla_{\boldsymbol{z}}a_R(\boldsymbol{z})}
{p_{\theta}(\boldsymbol{z}\mid\mathcal{D}_{\mathrm{ic}})+\delta}.
\end{equation}
The update in Equation~\ref{eq_6} subtracts the transformed raw gradient defined in Equation~\ref{eq_11} and adds Gaussian noise. The threshold $P=0.5$ follows the conventional binary-probability decision rule and was applied to every problem. Figures~\ref{fig_2} and~\ref{fig_3} show the gradient magnitudes and vector fields before the step transformation and noise addition.

\subsection{Candidate Initialization by Interpolation}
Candidate trajectories also depend on their starting points. Direct initialization starts from an observed Class~1 design. Interpolated initialization starts on the line segment between two Class~1 designs. GenPFN uses the interpolated seed before applying the initial perturbation in Equation~\ref{eq_5}.

For each candidate, GenPFN independently selects normalized Class~1 context vectors $\boldsymbol{z}_a$ and $\boldsymbol{z}_b$ and draws $r\sim\mathcal{U}(0,1)$. Equation~\ref{eq_12} defines the resulting seed:
\begin{equation}\label{eq_12}
\boldsymbol{z}_{\mathrm{seed}}
=
\boldsymbol{z}_a+r(\boldsymbol{z}_b-\boldsymbol{z}_a).
\end{equation}
The seed lies on the line segment joining the two observed designs. The initial perturbation in Equation~\ref{eq_5} and the updates in Equation~\ref{eq_6} allow the candidate to move beyond this segment. The configuration study crosses interpolated and direct Class~1 initialization with no guidance, continuous log-activated guidance, and ReLU-activated guidance to assess these design choices.

\subsection{Candidate Scaling and Evaluation}
After generation, inverse scaling returns each candidate to physical coordinates using the predefined parameter ranges. Problem-specific constraints assign feasible designs to Class~1 and all others to Class~0, while objective evaluation supplies the regression responses. Evaluation applies parameter scaling and discrete-variable reconstruction. Domain exits, constraint feasibility, objective validity, and Hull geometry success are recorded separately. Regression selects constraint-feasible candidates as described below.

\section{Experimental Methods}
The experiments assess the designs proposed by each method and their value for regression training. Feasibility, realism, coverage, and diversity describe the complete candidate pools. Regression then tests whether adding constraint-feasible candidates with externally evaluated responses improves prediction over training on the original feasible context alone.

GenPFN v1 and v3 use TabPFN v1 and v3 classifiers, respectively. The fourteen-problem benchmark varies context size and Class~0/Class~1 ratio to evaluate candidate quality and regression augmentation. Two Circles and Hull provide the primary examples, and the Appendix reports twelve additional problems. The PFN comparison evaluates three gradient activations under matched settings within each problem. SMOTE, TVAE, TabDDPM, and CTGAN supply conventional comparisons from matching context conditions. The eleven-problem complete factorial analysis examines initialization and guidance, and the four-problem long-run study follows their effects beyond the main 100-update horizon. The stern case demonstrates acquisition for an additional geometric requirement. Appendix~\ref{sec:software_environments} records the software environments and implementation settings.

\subsection{Test Problems}
Two Circles and Hull test candidate generation at different levels of complexity. Two Circles provides a visual two-parameter example with disconnected target regions. Hull tests a constrained 44-parameter engineering representation.

Two Circles defines Class~1 through two diagonally offset circular regions. The disconnected regions reveal whether candidates remain near observed designs, cover both target regions, or fall in the infeasible space between them. Regression uses the first objective of the KNO1 benchmark, a two-parameter, two-objective function with nonlinear and sinusoidal interactions~\citep{knowles2006parego}.

Hull uses a 44-parameter ship-hull representation with strong correlations among parameter subsets~\citep{bagazinski2023shipD}. GenPFN uses $\sigma=10^{-4}$ for Hull. The twelve supplementary problems extend the evaluation to other engineering constraints and design representations.

\subsection{Evaluation Metrics}
The evaluation first measures how many candidates satisfy Class~1, then describes their locations and spread. Realism measures how close generated candidates lie to reference designs. Coverage measures how closely the generated set reaches the reference designs. Diversity measures separation among generated candidates. Lower realism and coverage distances and higher diversity are preferred. All 5,000 proposals, including infeasible designs, enter these metrics, which are interpreted alongside class satisfaction. Regression uses constraint-feasible designs and their recorded evaluated responses. The supplementary problems and configuration study use the same definitions.

\subsubsection{Class Satisfaction Rate (Feasibility)}
Feasibility, or class satisfaction rate, is the fraction of candidates satisfying the ground-truth constraints that define Class~1. A higher rate means that more evaluations produce designs in the desired class.

\subsubsection{Realism and Coverage Chamfer Distance}
The context supplies labeled examples to the classifier. A fixed set of 20,000 Class~1 reference designs measures candidate realism and coverage and supplies the regression-validation targets. Figures labeled ``Large Prior'' refer to this reference set. Hull's positive context and reference sets partition an existing design library. For Heat Exchanger, coverage is interpreted relative to the fixed reference set rather than as uniform coverage of the feasible domain.

Equation~\ref{eq_13} defines the directed Chamfer distance (CD) from set $A$ to set $B$ as the mean Euclidean distance from each member of $A$ to its nearest neighbor in $B$:

\begin{equation}
CD = \frac{\sum^{N}_{i=1} \Vert A_i - B_{i*} \Vert}{N}
\label{eq_13}
\end{equation}

Here, $B_{i*}$ is the nearest neighbor of $A_i$. Let $G$ denote the 5,000 candidates and $V$ the 20,000-design Class~1 reference set. Realism measures the directed CD from $G$ to $V$, and coverage measures the directed CD from $V$ to $G$~\citep{regenwetter2023beyond}. Lower realism means that candidates lie closer to feasible reference designs. Lower coverage distance means that the candidate pool reaches more of the reference domain.

Realism and coverage are the two directed distances; their sum is bidirectional CD. Chamfer and MNN distances use range-normalized coordinates, while constraints and objectives use inverse-scaled physical coordinates. Distance comparisons are made within each problem because dimensionality and feasible-domain geometry differ across problems.

\subsubsection{Mean Nearest-Neighbor Distance (Within-Set Dispersion)}
Mean nearest-neighbor (MNN) distance measures dispersion within the candidate pool. For $G=\{\boldsymbol{g}_i\}_{i=1}^{N}$, Equation~\ref{eq_14} gives the mean distance from each candidate to its nearest other candidate:
\begin{equation}\label{eq_14}
D_{\mathrm{MNN}}(G)
=
\frac{1}{N}\sum_{i=1}^{N}
\min_{\substack{1\leq j\leq N\\j\neq i}}
\left\lVert\boldsymbol{g}_i-\boldsymbol{g}_j\right\rVert_2.
\end{equation}
Each candidate's nearest-neighbor search excludes its own index, while distinct duplicate rows remain zero-distance neighbors. A larger MNN therefore means that candidates are more widely separated~\citep{regenwetter2023beyond}. The tables report this range-normalized metric as \emph{diversity}.

\subsubsection{Regression Validation Performance}
The regression analysis tests the value of the acquired designs for performance prediction. AutoGluon is trained on the feasible context alone and on that context augmented with constraint-feasible candidates and their externally evaluated responses.

Candidate generation and distance metrics use parameter-range-normalized coordinates, while engineering constraints and objectives are evaluated in physical coordinates. Regression inputs and responses are standardized using the training-set means and standard deviations. The same fitted input scaler transforms validation designs, and predictions return to the original objective units before MSE is calculated.

Every sampling method uses the same AutoGluon workflow: \texttt{TabularPredictor} with \texttt{problem\_type='regression'}, a 15-second fit limit, parallel fitting on the allocated processor cores, and otherwise default settings. AutoGluon uses its version-dependent default regression model-selection procedure, with no explicit fitting seed, preset, or selection metric supplied.

The study evaluates the resulting predictions using mean squared error (MSE) in the original objective units. Context-only fitting requires more than four feasible rows. Zero-yield pools and failed fits retain missing augmented MSE, without a context-only fallback.

Regression training retains constraint-feasible candidates and their evaluated responses.

\subsection{Experimental Procedure}
The saved Class~0 and Class~1 libraries supply the labeled context, and stratification sets its class ratio. The setting therefore starts with known examples of both classes. A context condition specifies the total number of designs and their Class~0/Class~1 ratio. Each condition contains ten labeled context sets, identified by subset IDs 0--9. The same saved Class~1 reference set supplies distance and regression validation within each problem. Each problem follows the same generate--evaluate--augment procedure:

\begin{enumerate}
    \item Draw a saved context with a specified total size and Class~0/Class~1 ratio. The nominal ratios are 20/80, 50/50, and 80/20. At total size four, these contain 1/3, 2/2, and 3/1 designs, respectively.
    \item Provide each method with the same Class~0 and Class~1 context designs and their labels as GenPFN.
    \item Generate 5,000 candidates targeted toward Class~1. PFN configurations process all 5,000 queries together and return the final state after 100 updates.
    \item Evaluate every candidate with the ground-truth formulation, and calculate feasibility, realism, coverage, and diversity using all generated candidates.
    \item Fit AutoGluon on the feasible context alone and on the same context augmented with constraint-feasible generated designs and their evaluated responses. Evaluate MSE on the validation set. Zero-feasible yield and failed fits remain distinct missing outcomes.
    \item Repeat the procedure for all three ratios; total context sizes 4, 16, 64, 128, 256, and 1,024; and saved subset IDs 0--9, giving ten contexts per size--ratio condition.
\end{enumerate}

The three ratios, six context sizes, and ten repetitions give 180 intended contexts per problem and 2,520 across fourteen problems. Comparisons match problem, context size, class ratio, and subset identifiers. Small-context SMOTE conditions without a candidate pool are recorded as missing.

Each regression comparison uses the same budget of 5,000 proposals to evaluate the complete augmentation workflow. A method's constraint-feasible yield determines how many designs it adds to training. Each augmented fit is paired with its recorded context-only reference.

Pure interpolation stops at the Class~1 interpolation seed. Interpolation plus noise applies Gaussian perturbations without classifier guidance, using matched starts and the main-run noise scale. GenPFN adds gated guidance to this interpolated, noisy trajectory. The PFN configurations retain the labeled context, row order, interpolated starts, paired random streams within each condition, and common coefficients within each problem. Local TabPFGen uses continuous stabilized probability guidance. The TabEBM activation differentiates $-\log\sum_c\exp s_c$ on the real context. These labels denote controlled activations in the common sampler, rather than complete reproductions of the published generators.

For v3 energy guidance, the recorded adapter uses $\widetilde{s}_c=s_c^{\mathrm{native}}+r$, with $r=\operatorname{mean}_h\log\sum_j\exp a_{hj}$ and $a_{hj}$ the query--key attention score for head $h$ and context row $j$, including learned query scaling. This class-shared offset preserves normalized probabilities and changes the energy gradient; it does not recover a native v1-style head. Both probability-guidance methods use native probabilities.

Sampling coefficients are fixed within each problem across activations and backbones. Two Circles and Ring use $\alpha=0.1$ and $\lambda=2$. Heat Exchanger uses $\alpha=10$ and $\lambda=0.0002742$; Compression Spring uses $\alpha=10$ and $\lambda=0.0001596$; Hull uses $\alpha=1$ and $\lambda=0.0001319$. The other nine problems use $\alpha=0.001$ and $\lambda=0.1$. These settings are author-selected rather than the result of systematic hyperparameter optimization. Displayed values use at most four significant figures; full-precision coefficients and results are retained with the supporting data.

The noise scale remains $\sigma=5\times10^{-4}$ except Hull, which uses $10^{-4}$. For the noise-only control, initialization plus 100 updates contribute 101 independent Gaussian draws, giving per-coordinate displacement standard deviation $\sqrt{101}\sigma$. The noise is independent of the deterministic cap $\alpha\lambda$.

\subsection{Hierarchical Paired Statistical Analysis}
The analysis first compares methods on matching context sets, identified by problem, context size, Class~0/Class~1 ratio, and subset. Repetitions are averaged within each context-size and class-ratio condition, and the conditions are summarized within each engineering problem. A random-effects model then combines the fourteen problem-level estimates. Each engineering problem remains the unit of evidence in the pooled comparison.

In the main GenPFN comparisons, positive feasibility differences favor GenPFN. Realism, coverage, and MSE are expressed as comparator/GenPFN ratios, while diversity uses GenPFN/comparator; ratios above one favor GenPFN. In the general paired definitions, $L$ denotes the named method and $R$ its comparator. Feasibility uses $100(f_L-f_R)$ in percentage points (pp). Realism, coverage, and MSE use $\log(R/L)$, while diversity uses $\log[(L+\epsilon)/(R+\epsilon)]$. The logarithm places proportional increases and decreases on a symmetric scale. Exponentiating a pooled log estimate gives a geometric mean ratio, rather than a ratio of pooled raw means.

A diversity value of zero requires a correction before taking logarithms. If a problem--comparison contains zero, $\epsilon$ is half its smallest positive paired diversity value; otherwise, $\epsilon=0$. Pure interpolation can have zero diversity when only one Class~1 seed is available. Additional analyses exclude zero-diversity pairs or compare untransformed diversity differences to examine whether the conclusions depend on that correction.

Conditions with at least two matching pairs receive equal weight within each problem. Let $J_p$ be the number of retained conditions for problem $p$, $n_{pc}$ the number of paired repetitions in condition $c$, and $s^2_{pc}$ the sample variance of their effects. The within-problem sampling variance is $J_p^{-2}\sum_c s^2_{pc}/n_{pc}$. Because contexts at different sizes can share source rows, this sampling-variance calculation is approximate.

Primary MSE comparisons use contexts containing at least 16 designs, with an observed context-only fit and positive finite MSE for both methods being compared. Each retained condition requires at least two matching pairs. Unavailable pools, zero-feasible yields, and failed fits remain missing outcomes. Regression availability is reported alongside MSE for these eligible pairs. A common-support sensitivity restricts conventional comparisons to conditions with results for all four baselines, and an equal-problem sensitivity avoids precision weighting.

Confidence intervals (CIs) describe uncertainty in the average comparison across problems. Prediction intervals describe the range of effects allowed by the random-effects model for another problem. The between-problem standard deviation, $\tau$, measures heterogeneity.

The analysis combines the fourteen problem estimates using an intercept-only random-effects model fitted by restricted maximum likelihood (REML). All identified stationary points and the zero boundary are compared to avoid choosing a local likelihood optimum. A recorded numerical floor stabilizes very small or zero sampling variances. Modified Hartung--Knapp inference bounds its variance-scale factor below by one~\citep{hartung2001overall}. Pointwise 95\% CIs use $t_{k-1}$ for $k$ problems; prediction intervals use $t_{k-2}\sqrt{\tau^2+\mathrm{SE}^2}$. These intervals are not adjusted for multiplicity. The normal random-effects approximation is unbounded, so feasibility intervals are reported without clipping.

Significance statements use the declared adjusted tests. A comparison family groups related questions, such as comparisons of GenPFN v1 with the conventional generators. For each main endpoint and question family, a maximum-$|t|$ sign-flip test enumerates problem-level sign patterns under joint sign symmetry, holds REML weights and heterogeneity fixed, and recomputes Hartung--Knapp studentization~\citep{westfall1993resampling}. Holm adjustment then covers endpoint--contrast tests within that family and population~\citep{holm1979sequential}. The families are conventional comparisons within each backbone, activation comparisons within each backbone, backbone comparisons, shared-control comparisons, and augmentation versus context-only MSE. Primary all-proposal tests use fourteen problems. The reported intervals describe individual effects without adjustment for multiple comparisons. An interval excluding the null can therefore accompany a nonsignificant adjusted $p$ value. HK-$t$/Holm results provide a sensitivity. The analysis is exploratory because its families were chosen after the results were available; no single familywise guarantee covers all supplementary comparisons.

\subsection{Secondary Factorial Configuration Analysis}
A secondary factorial analysis examines how candidate initialization and gradient guidance affect feasibility. The TabPFN v1 comparison crosses direct and interpolated initialization with no, gated, and continuous probability guidance. Appendix~\ref{sec:ablation} provides the configurations, paired analysis procedure, and results.

\begin{figure}[t]
\centering\includegraphics[width=2.75in]{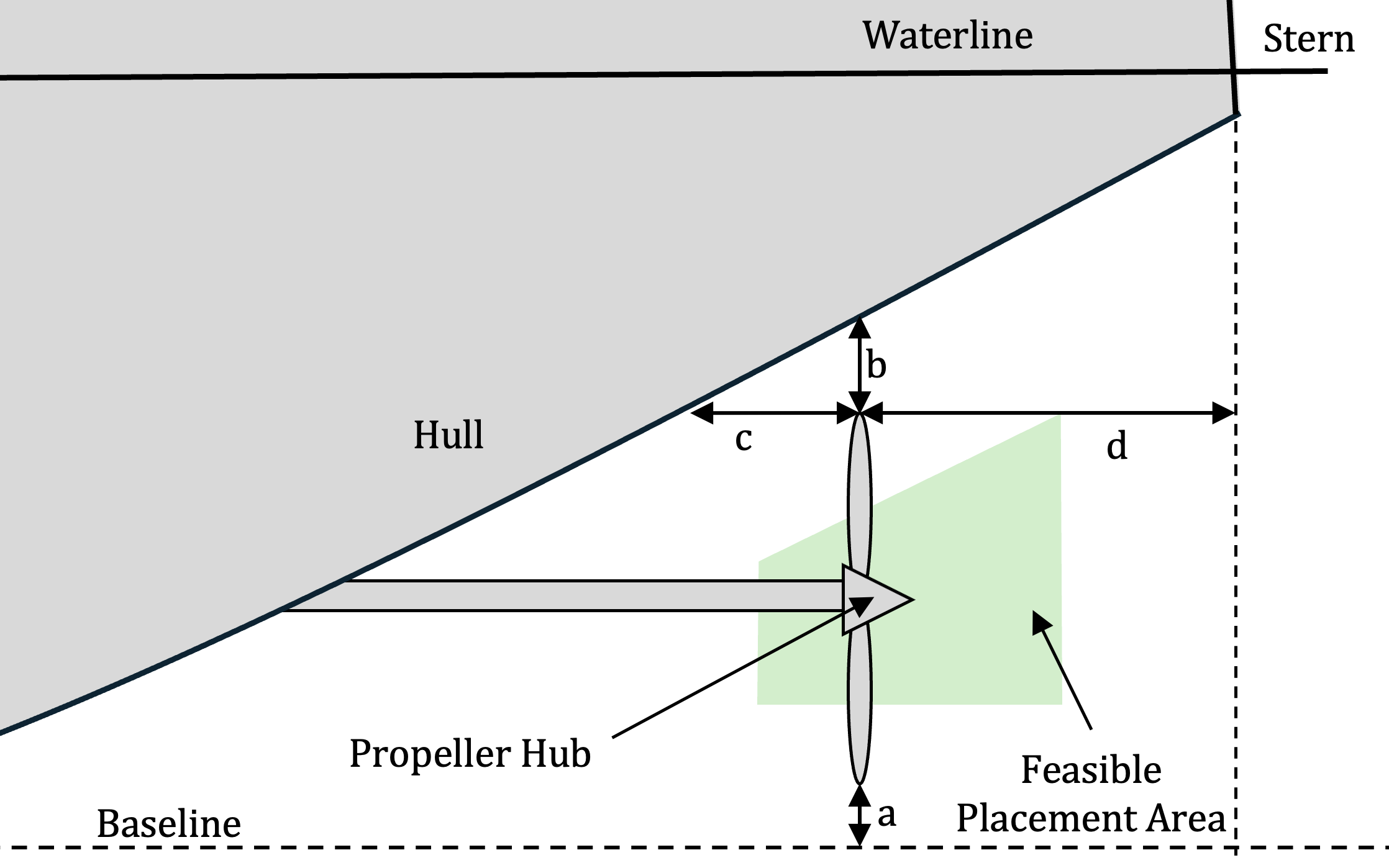}
\caption{Required clearances for the 3.2-m-diameter propeller: $a>32$~mm, $b>640$~mm, $c>960$~mm, and $d>1{,}600$~mm.}
\label{fig_4}
\end{figure}

\begin{figure}[t]
\centering\includegraphics[width=3.0in]{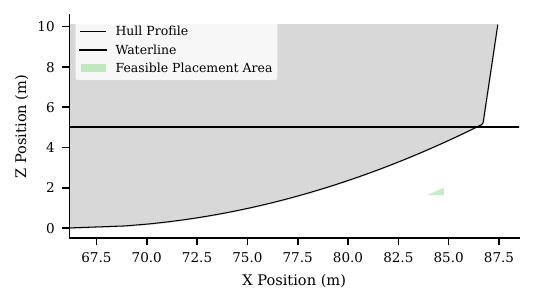}
\caption{Row-17 parent hull with a $0.1519\,\mathrm{m}^2$ admissible placement area for the propeller-hub center.}
\label{fig_5}
\end{figure}

\begin{figure}[t]
\centering\includegraphics[width=0.95\linewidth]{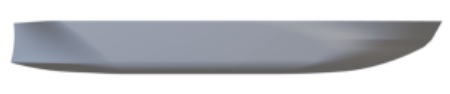}
\caption{Initial hull for the stern-design case study.}
\label{fig_6}
\end{figure}
\subsection{Design Study: Hull Stern Inpainting}
The hull-stern study applies GenPFN when an existing design must satisfy an additional geometric requirement. Previous generative hull-design studies produced designs with less drag than designs in the training data and prior optimization studies~\citep{bagazinski2023shipgen,bagazinski2024cshipgen}. Ship-D labels describe bare-hull drag, so compatibility with a conventional propeller requires a separate geometric evaluation. Design inpainting varies the stern while keeping the bow and parallel midbody fixed. This allows the study to acquire propeller-compatible stern alternatives while preserving the rest of the hull.

The case hull is zero-based row~17 of \texttt{CSOV\_X\_Mod.csv}, with an overall length of 87.44~m, a 5.0-m draft, and a 3.2-m-diameter propeller. The same parent defines the fixed controls, sampling-window centers, and displayed baseline. Figure~\ref{fig_4} defines the required clearances. Figure~\ref{fig_5} shows its $0.1519\,\mathrm{m}^2$ admissible propeller-hub placement area. A larger admissible region provides more flexibility in locating the hub during early design~\citep{page2012flexibility}. The study tests whether GenPFN can acquire stern designs with larger placement areas.

Figure~\ref{fig_6} shows the parent hull, which defines the geometry retained during inpainting. Random sampling varies eighteen stern controls, including one binary control, within prescribed windows around the parent. The remaining twenty-seven stored coordinates stay fixed. This representation stores 45 coordinates, including overall length. Hull preprocessing includes geometric cleanup and binary-control reconstruction.

Geometric evaluation checks hull feasibility and calculates the admissible propeller-placement region. Class~1 requires a feasible hull, a feasible propeller-placement region, and area strictly greater than $0.5\,\mathrm{m}^2$; all other context designs receive Class~0.

All 2,000 labeled designs form the TabPFN v1 context with two ensemble configurations. GenPFN proposes 1,000 candidates in one batch from interpolated Class~1 starts, using per-candidate L2 clipping, $\alpha=0.001$, $\lambda=0.1$, $\sigma=0.0005$, and 100 updates. The same geometric procedure evaluates every candidate, and the stored coordinates outside the mutable stern block are checked against the parent.

\begin{figure*}[t]
\centering\includegraphics[width=6.0in]{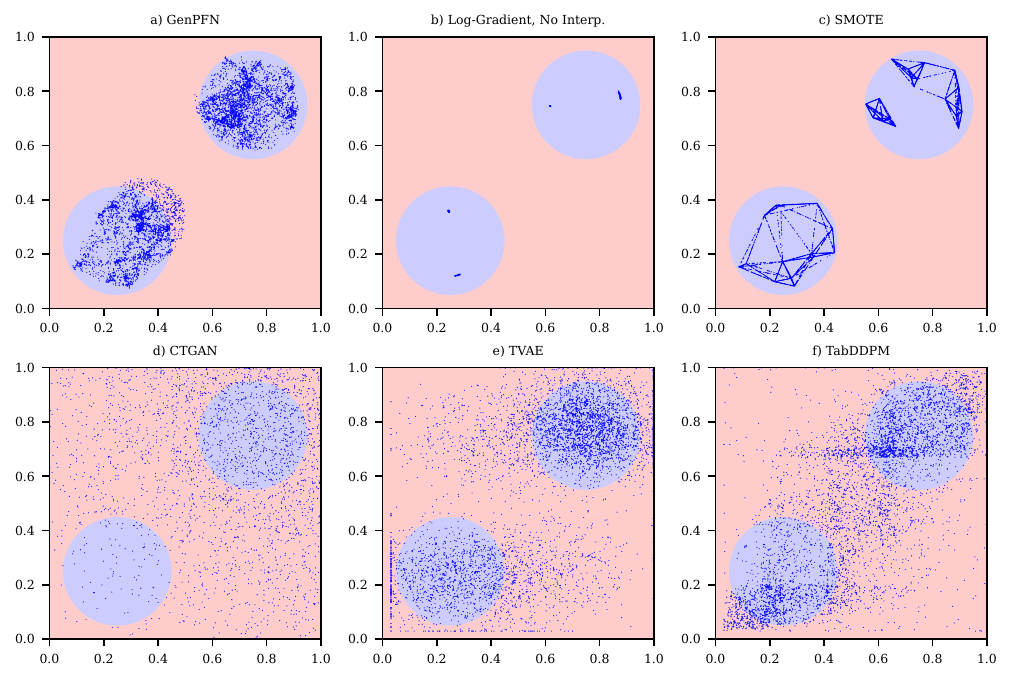}
\caption{Candidate distributions for the \emph{Two Circles} problem using one 64-design context with equal class proportions and subset 0: (a) GenPFN v1, (b) continuous log-activated guidance with direct initialization, (c) SMOTE, (d) CTGAN, (e) TVAE, and (f) TabDDPM. Blue markers denote proposals, while the background gives the ground-truth class regions. Axes show the original unit-square viewport; proposals outside this viewport remain part of the evaluated pool. CTGAN's mean Class~1 rate of 17.8\% summarizes the full Two Circles benchmark.}
\label{fig_7}
\end{figure*}

\section{GenPFN Sampling Results}
The results first examine how often candidates reach Class~1 and how widely they spread through the design space. The regression analysis then tests whether the verified designs improve training. Two Circles and Hull provide the primary examples, and the paired hierarchical analysis combines comparisons across all fourteen selected problems. The Appendix extends these results with configuration studies and the long-run noise experiment.

\begin{figure*}[t]
\centering\includegraphics[width=6.5in]{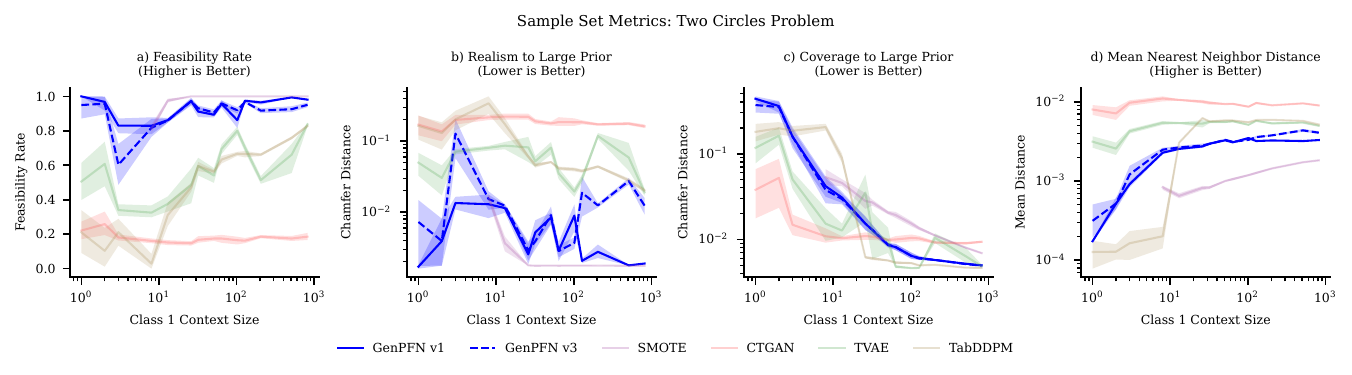}
\caption{Sample-set metrics for Two Circles versus the number of \emph{Class~1} context samples: (a) feasibility, (b) realism Chamfer distance, (c) coverage Chamfer distance, and (d) mean nearest-neighbor distance. Solid and dashed blue lines distinguish GenPFN v1 and v3; the remaining methods retain the original colors. Higher feasibility and nearest-neighbor distance and lower realism and coverage distances are preferred. Lines show means; bands show pointwise 95\% context-bootstrap intervals.}
\label{fig_8}
\end{figure*}

\begin{figure*}[t]
\centering\includegraphics[width=6.5in]{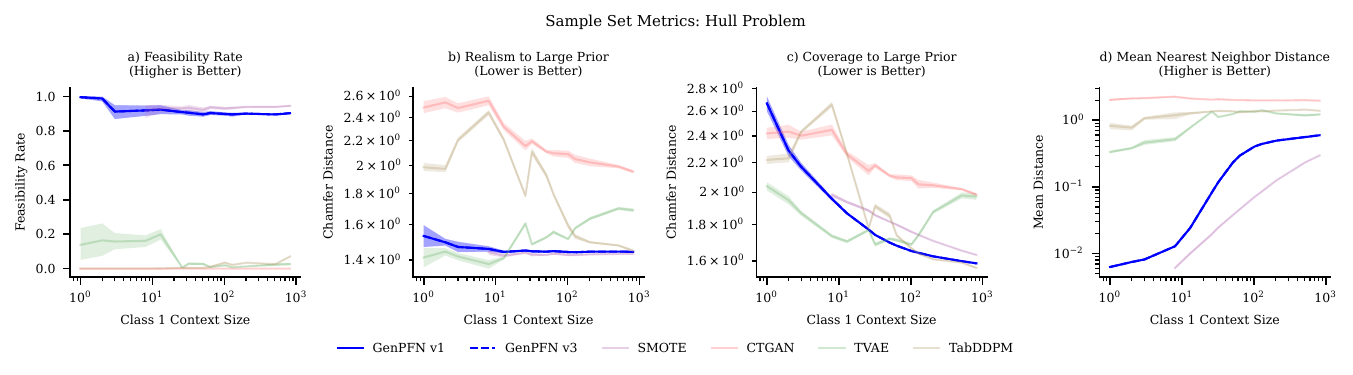}
\caption{Sample-set metrics for Hull versus the number of \emph{Class~1} context samples, using the same panels, method colors, backbone line styles and pointwise 95\% context-bootstrap intervals as Figure~\ref{fig_8}.}
\label{fig_9}
\end{figure*}

\subsection{Sample Set Metrics}
Figure~\ref{fig_7} shows candidate placement in Two Circles, and Figures~\ref{fig_8} and~\ref{fig_9} quantify feasibility, realism, coverage and diversity for Two Circles and Hull. The horizontal axis counts Class~1 context designs rather than total context size. Different combinations of total size and class ratio can therefore contribute to the same position.

Curves average conditions sharing a Class~1 count, and shaded bands show pointwise 95\% context-bootstrap intervals. Conditions are first averaged within each context subset; the ten subset IDs are then resampled using the recorded deterministic 2,000-draw calculation. The regression plots use the same Class~1 context count on their horizontal axes. The paired hierarchical analysis separately compares methods across problems.

The Two Circles example reveals the balance between spread and feasibility. SMOTE follows local interpolation paths, while GenPFN spreads candidates within the target regions. Continuous guidance with direct starts concentrates this example near a few locations. CTGAN, TVAE and TabDDPM also place candidates outside Class~1. Whole-pool spread and class satisfaction therefore describe different aspects of the proposed designs.

\begin{table}[h]
\caption{Mean sample-set metrics over the evaluated conditions for Two Circles and Hull. Both GenPFN backbones and the conventional generators propose 5,000 candidates per available context. Every proposal enters the geometry metrics. SMOTE lacks part of the grid, so its raw means have different support; paired inference uses matched conditions. Bold denotes the best displayed value per problem and metric, including ties.}
\label{tab_1}%
\centering{%
\fitTableToWidth{\columnwidth}{
\begin{tabular}{cl|cccc}
\toprule
 && \multicolumn{4}{c}{Sample Set Metrics}\\
Problem & Method & Feasibility ($\uparrow$) & Realism ($\downarrow$) & Coverage ($\downarrow$) & Diversity ($\uparrow$) \\
\midrule
\multirow{6}{*}{Hull} & GenPFN v1 & 0.9150 & 1.457 & 1.851 & 0.2245 \\
 & GenPFN v3 & 0.9150 & 1.457 & 1.851 & 0.2249 \\
 & SMOTE & \textbf{0.9318} & \textbf{1.431} & \textbf{1.802} & 0.0805 \\
 & CTGAN & $<0.0001$ & 2.222 & 2.195 & \textbf{2.055} \\
 & TVAE & 0.0764 & 1.520 & 1.811 & 0.9846 \\
 & TabDDPM & 0.0141 & 1.879 & 1.962 & 1.256 \\
\midrule
\multirow{6}{1cm}{Two \newline Circles} & GenPFN v1 & 0.9195 & 0.0064 & 0.0725 & 0.0025 \\
 & GenPFN v3 & 0.8847 & 0.0234 & 0.0678 & 0.0027 \\
 & SMOTE & \textbf{0.9831} & \textbf{0.0031} & 0.0233 & 0.0011 \\
 & CTGAN & 0.1780 & 0.1870 & \textbf{0.0145} & \textbf{0.0094} \\
 & TVAE & 0.5421 & 0.0650 & 0.0296 & 0.0050 \\
 & TabDDPM & 0.4700 & 0.1028 & 0.0661 & 0.0039 \\
\bottomrule
\end{tabular}
}%
}
\end{table}

Table~\ref{tab_1} reports descriptive means for Two Circles and Hull. The paired analysis combines comparisons across all fourteen problems. GenPFN v1 produces higher Class~1 rates and lower realism distances than the tested neural generators. Its class satisfaction exceeds CTGAN, TVAE and TabDDPM by 69.54, 36.90 and 44.25 percentage points, respectively; the adjusted feasibility comparisons are significant. TVAE provides lower coverage distance and higher diversity, while CTGAN also provides higher diversity. The adjusted coverage comparison with CTGAN and the coverage and diversity comparisons with TabDDPM do not establish differences.

Relative to SMOTE, GenPFN expands coverage and diversity beyond local interpolation. The realism comparison does not establish a difference. Its class-satisfaction difference is -5.79 percentage points, with adjusted $p=0.2545$. Candidate feasibility, coverage and dispersion therefore favor different aspects of acquisition.

The supporting data report whether the diversity findings persist when zero-diversity pairs are excluded or untransformed differences are used. Appendix~\ref{sec:ablation} examines initialization and guidance using the factorial results.

Appendix~\ref{sec:pfn_sensitivity} compares GenPFN with interpolation and interpolation plus noise on matching contexts. At 100 updates, the pooled GenPFN-v1 feasibility difference from noisy interpolation is 1.61 percentage points (pointwise 95\% CI: -0.82 to 4.04). The result does not establish a suite-wide advantage under the adjusted test. Appendix~\ref{app:long_noise} separately examines guidance and cumulative noise over longer trajectories.

\begin{figure}[t]
\centering\includegraphics[width=2.5in]{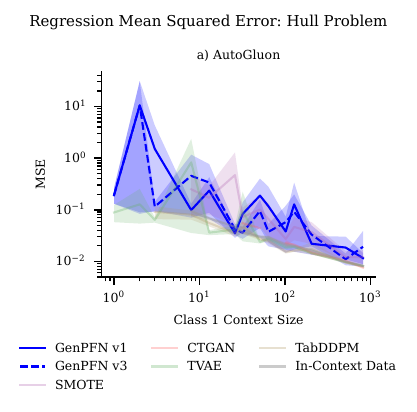}
\caption{Hull AutoGluon validation $MSE$ versus the number of \emph{Class~1} context samples. Only externally evaluated feasible candidates augment the context. Missing fits remain gaps.}
\label{fig_10}
\end{figure}

\begin{figure}[t]
\centering\includegraphics[width=2.5in]{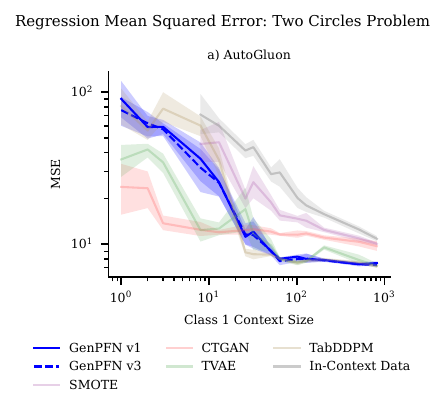}
\caption{Two Circles AutoGluon validation $MSE$ versus the number of \emph{Class~1} context samples. The context-only curve uses the same reference fits as the paired analysis.}
\label{fig_11}
\end{figure}
\subsection{Regression Results}
Class satisfaction determines which generated designs enter regression training. Each method proposes 5,000 candidates, and external evaluation supplies feasibility labels and objective values. Verified candidates augment the original feasible context, so training-set size varies with feasibility yield. Validation on the fixed 20,000-design Class~1 reference measures the complete generate--evaluate--augment workflow.

Validation on Class~1 designs measures predictive accuracy within the chosen region of interest. Improvements in this accuracy quantify how the acquired observations support learning about designs with the desired traits.

Figures~\ref{fig_10} and~\ref{fig_11} show AutoGluon validation mean squared error against the number of Class~1 context designs.

Table~\ref{tab_2} reports descriptive means for Two Circles and Hull, using successful fits with an observed context-only reference and total context size at least 16. Values are comparable within a problem. The pooled regression improvement comes from the paired log-effect analysis across fourteen problems, rather than division of the arithmetic means shown for these two examples.

\begin{table}[h]
\caption{Mean observed successful AutoGluon validation $MSE$ at total context sizes 16--1,024 with an observed context-only fit. Each method's available successful fits determine its descriptive mean; missing augmented fits are not imputed. Augmentation adds constraint-feasible proposals and their evaluated responses. Eligible regression-pair counts are reported separately in Table~\ref{tab_3}. Bold denotes the lowest displayed mean MSE within each problem, including ties.}
\label{tab_2}%
\centering{%
\fitTableToWidth{\columnwidth}{%
\begin{tabular}{ll|c}
\toprule
Problem & Method & Mean Squared Error ($\downarrow$)\\
\midrule
\multirow{7}{*}{Hull} & GenPFN v1 & 1.011E-01 \\
 & GenPFN v3 & 1.195E-01 \\
 & In-context only & 3.522E-02 \\
 & SMOTE & 1.120E-01 \\
 & CTGAN & \textbf{2.531E-02} \\
 & TVAE & 8.503E-02 \\
 & TabDDPM & 2.582E-02 \\
\midrule
\multirow{7}{1cm}{Two \newline Circles} & GenPFN v1 & 1.313E+01 \\
 & GenPFN v3 & 1.270E+01 \\
 & In-context only & 3.262E+01 \\
 & SMOTE & 2.234E+01 \\
 & CTGAN & 1.161E+01 \\
 & TVAE & \textbf{1.025E+01} \\
 & TabDDPM & 1.446E+01 \\
\bottomrule
\end{tabular}
}
}
\end{table}

Table~\ref{tab_3} traces the displayed methods from intended contexts to available candidate pools and successful augmented fits. A candidate pool can be available even when it produces no feasible designs or no successful regression fit. A paired comparison additionally needs the corresponding context-only fit and must satisfy the eligibility rules in Experimental Methods. No zero-yield pool receives a context-only fallback, and no failed or unavailable result is imputed.

\begin{table*}[t]
\caption{Candidate-pool and AutoGluon availability for the six displayed methods. Intended-context, available-pool, zero-feasible-pool, and observed-fit counts cover all 2,520 intended contexts per method. In-context observed counts are successful context-only fits. Positive-feasibility fit missing counts identify pools with feasible candidates but no observed augmented fit. Primary paired $N$ counts eligible method-versus-context-only pairs under the Methods criteria. No zero-yield or failed fit receives a fallback. Added rows are median [first--third quartile] among successful augmented fits, using each fit's saved generated-training-row count. These descriptive yields are not restricted to primary pairs.}
\label{tab_3}
\centering
\fitTableToWidth{\textwidth}{%
\begin{tabular}{lrrrrrrrrl}
\toprule
Method & \shortstack{Intended\\contexts} & \shortstack{In-context\\observed} & \shortstack{Available\\pools} & \shortstack{Zero-feasible\\pools} & \shortstack{Observed\\fits} & \shortstack{Positive-feasibility\\fit missing} & \shortstack{In-context\\fallbacks} & \shortstack{Primary paired\\$N$} & \shortstack{Feasible rows added\\median [first--third quartile]} \\
\midrule
GenPFN v1 & 2,520 & 1,960 & 2,520 & 0 & 2,520 & 0 & 0 & 1,960 & 4,987 [4,532--4,999] \\
GenPFN v3 & 2,520 & 1,960 & 2,520 & 0 & 2,519 & 1 & 0 & 1,960 & 4,986 [4,531--4,999] \\
SMOTE & 2,520 & 1,960 & 1,960 & 0 & 1,960 & 0 & 0 & 1,960 & 5,000 [5,000--5,000] \\
CTGAN & 2,520 & 1,960 & 2,520 & 172 & 2,346 & 2 & 0 & 1,826 & 1,028 [428.3--1,913] \\
TVAE & 2,520 & 1,960 & 2,520 & 1 & 2,518 & 1 & 0 & 1,960 & 3,246 [1,623--4,119] \\
TabDDPM & 2,520 & 1,960 & 2,520 & 172 & 2,341 & 7 & 0 & 1,892 & 3,018 [1,508--3,894] \\
\bottomrule
\end{tabular}%
}
\end{table*}

The added feasible-row counts describe how much evaluated data each successful fit receives. Regression performance therefore reflects both the amount acquired and its value for training. Table~\ref{tab_4} combines the paired sample-set and regression estimates across fourteen problems.

\begin{table*}[t]
\caption{Paired random-effects results for GenPFN v1 across fourteen engineering problems. The model combines one estimate per problem; $N$ counts retained context pairs. Ratios above one favor GenPFN: comparator/GenPFN for realism, coverage, and MSE, and GenPFN/comparator for diversity. Feasibility is GenPFN minus comparator in pp. Modified Hartung--Knapp 95\% CIs and prediction intervals are pointwise and unadjusted for multiplicity. Between-problem standard deviation $\tau$ uses pp for feasibility and log ratios otherwise. Adjusted $p$ values follow the declared maximum-$|t|$/Holm families. Regression retains supported successful pairs. Bold adjusted $p$ values denote statistically significant comparisons.}
\label{tab_4}
\centering
\fitTableToWidth{\textwidth}{%
\begin{tabular}{llccccc}
\toprule
Endpoint & Comparator & Oriented effect [95\% CI] & $\tau$ & 95\% prediction interval & $N$ & $p_{\mathrm{adj}}$ \\
\midrule
Feasibility & SMOTE & $-5.79 [-11.92, 0.35] pp$ & 10.62 & $[-29.73, 18.16] pp$ & 1,960 & $0.2545$ \\
Feasibility & CTGAN & $69.54 [61.08, 78.01] pp$ & 14.64 & $[36.53, 102.6] pp$ & 2,520 & $\mathbf{0.0024}$ \\
Feasibility & TVAE & $36.90 [26.25, 47.56] pp$ & 18.44 & $[-4.69, 78.50] pp$ & 2,520 & $\mathbf{0.0024}$ \\
Feasibility & TabDDPM & $44.25 [35.09, 53.42] pp$ & 15.86 & $[8.49, 80.02] pp$ & 2,520 & $\mathbf{0.0024}$ \\
\midrule
Realism & SMOTE & $0.847 [0.643, 1.116]$ & 0.476 & $[0.290, 2.477]$ & 1,960 & $0.6140$ \\
Realism & CTGAN & $13.19 [5.580, 31.16]$ & 1.489 & $[0.459, 379.0]$ & 2,520 & $\mathbf{0.0024}$ \\
Realism & TVAE & $2.891 [1.701, 4.914]$ & 0.918 & $[0.365, 22.93]$ & 2,520 & $\mathbf{0.0146}$ \\
Realism & TabDDPM & $5.442 [2.783, 10.64]$ & 1.161 & $[0.397, 74.59]$ & 2,520 & $\mathbf{0.0033}$ \\
\midrule
Coverage & SMOTE & $1.513 [1.311, 1.746]$ & 0.248 & $[0.865, 2.646]$ & 1,960 & $\mathbf{0.0027}$ \\
Coverage & CTGAN & $0.750 [0.620, 0.908]$ & 0.330 & $[0.356, 1.578]$ & 2,520 & $0.0916$ \\
Coverage & TVAE & $0.723 [0.662, 0.790]$ & 0.151 & $[0.515, 1.017]$ & 2,520 & $\mathbf{0.0024}$ \\
Coverage & TabDDPM & $0.918 [0.861, 0.979]$ & 0.110 & $[0.716, 1.176]$ & 2,520 & $0.1519$ \\
\midrule
Diversity & SMOTE & $4.140 [3.402, 5.038]$ & 0.340 & $[1.922, 8.915]$ & 1,960 & $\mathbf{0.0024}$ \\
Diversity & CTGAN & $0.153 [0.115, 0.204]$ & 0.499 & $[0.050, 0.471]$ & 2,520 & $\mathbf{0.0024}$ \\
Diversity & TVAE & $0.328 [0.251, 0.429]$ & 0.465 & $[0.115, 0.935]$ & 2,520 & $\mathbf{0.0024}$ \\
Diversity & TabDDPM & $0.965 [0.586, 1.591]$ & 0.863 & $[0.138, 6.769]$ & 2,520 & $0.9965$ \\
\midrule
Regression $MSE$ & In-context only & $2.638 [1.932, 3.603]$ & 0.536 & $[0.787, 8.844]$ & 1,960 & $\mathbf{0.0037}$ \\
Regression $MSE$ & SMOTE & $2.329 [1.834, 2.958]$ & 0.410 & $[0.923, 5.876]$ & 1,960 & $\mathbf{0.0024}$ \\
Regression $MSE$ & CTGAN & $0.341 [0.174, 0.666]$ & 1.158 & $[0.025, 4.641]$ & 1,826 & $0.0835$ \\
Regression $MSE$ & TVAE & $0.366 [0.253, 0.531]$ & 0.639 & $[0.087, 1.550]$ & 1,960 & $\mathbf{0.0027}$ \\
Regression $MSE$ & TabDDPM & $0.407 [0.272, 0.609]$ & 0.696 & $[0.085, 1.955]$ & 1,892 & $\mathbf{0.0146}$ \\
\bottomrule
\end{tabular}%
}
\end{table*}

Among observed paired regression fits, adding evaluated GenPFN-v1 candidates reduces estimated AutoGluon $MSE$ by 62.1\% relative to context-only training (pointwise 95\% CI: 48.2--72.2\% lower; adjusted $p=0.00366$). The context-only/GenPFN ratio is 2.64 (CI: 1.93--3.60), based on 1,960 matching contexts.

GenPFN also has lower paired MSE than SMOTE, with a comparator/GenPFN ratio of 2.33. TVAE and TabDDPM have lower paired MSE than GenPFN. CTGAN's point estimate favors CTGAN, but its adjusted comparison is not significant. These estimates describe observed paired fits; Table~\ref{tab_3} reports their availability.

The random-effects estimates vary across problems, so Table~\ref{tab_4} reports prediction intervals alongside confidence intervals for the average effects. Unbounded feasibility prediction intervals are reported without clipping. Complete-support and equal-problem sensitivity results are provided in the supporting data.

The guidance comparison asks whether adding classifier gradients improves acquisition beyond interpolation and noise. The noise/GenPFN-v1 MSE ratio is 1.020 (pointwise 95\% CI: 0.999--1.041). The augmentation gain relative to context-only training therefore does not establish the same gain from classifier guidance alone.

The version comparison asks how changing the underlying TabPFN classifier affects the same generation procedure. The pooled v3-minus-v1 GenPFN feasibility difference is 0.0025 percentage points (CI: -0.0018 to 0.0068); this comparison does not establish a consistent advantage. The controlled PFN results remain in Appendix~\ref{sec:pfn_sensitivity}.

The benchmark uses a common budget of 5,000 candidate evaluations per method.

\subsection{Sampling Computational Effort}
The stern acquisition round generated 1,000 candidates using 100 updates and a context of 2,000 labeled designs. Candidate sampling took 11.97 seconds, and subsequent geometry evaluation took 0.80 seconds, as reported in Table~\ref{tab_5}.

\begin{table}[h]
\caption{Recorded computational effort for the single stern acquisition round: 1,000 candidates in one batch, 100 updates, and 2,000 labeled context designs. Sampler and generated-candidate geometry-evaluation times are separate observations from this workload; no cross-method runtime ranking is inferred.}
\label{tab_5}%
\centering{%
\fitTableToWidth{\columnwidth}{%
\begin{tabular}{@{}l|r@{}}
\toprule
Stage & Recorded Time [s] \\
\midrule
GenPFN candidate sampling & 11.97 \\
Generated-candidate geometry evaluation & 0.80 \\
\bottomrule
\end{tabular}
}%
}
\end{table}

\subsection{GenPFN Use Cases}
The sample-set and regression results show how the methods acquire designs. SMOTE keeps candidates close to known feasible examples, while the conventional neural generators can provide broader spread after task-specific training. GenPFN obtains a larger Class~1 fraction than the tested neural generators. External evaluation identifies the designs that expand the engineering dataset.

When engineering requirements change, the evaluated context can be relabeled to define a new desired class. Like SMOTE, GenPFN can reuse these examples without fitting a new task-specific generator. The same pretrained classifier can then guide candidate generation without changes to its weights. The stern application demonstrates acquisition for an additional propeller-compatibility requirement while preserving the rest of the parent hull.

\begin{figure}[t]
\centering\includegraphics[width=2.75in]{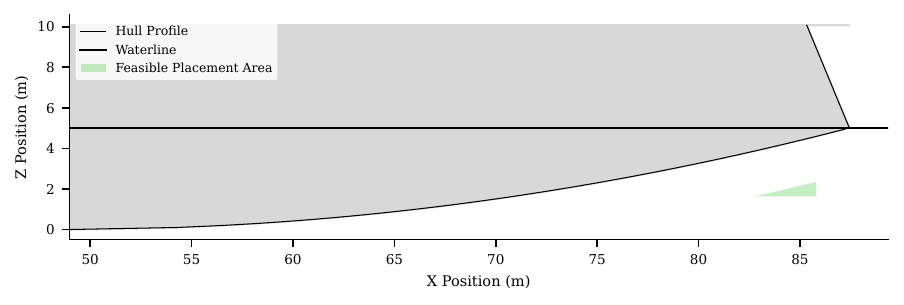}
\caption{Evaluated GenPFN stern candidate with a $1.079\,\mathrm{m}^2$ propeller-hub placement area.}
\label{fig_12}
\end{figure}

\begin{figure}[t]
\centering\includegraphics[width=02.75in]{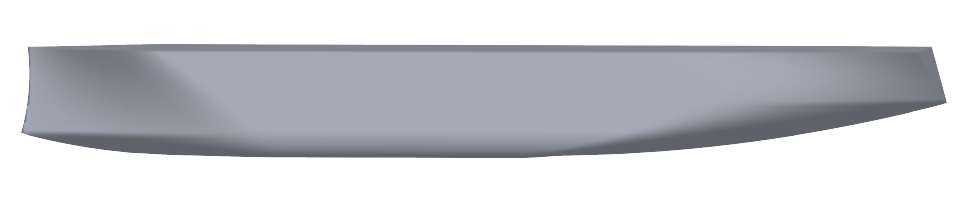}
\caption{Overall hull for the candidate in Figure~\ref{fig_12}; only the stern parameters differ from the initial hull in Figure~\ref{fig_6}.}
\label{fig_13}
\end{figure}
\section{Hull Stern Inpainting Proof of Concept}
The stern study applies class-targeted generation and external evaluation to propeller compatibility. Of 2,000 randomly sampled context designs, 28 (1.4\%) passed the geometric checks and provided a propeller-hub placement area greater than $0.5\,\mathrm{m}^2$. Among randomly sampled context designs with nonzero placement area, the mean area was $0.2319\,\mathrm{m}^2$. The desired class was sparsely represented in the context.

GenPFN used all 2,000 labeled designs as context and proposed 1,000 Class~1 candidates. The same geometric procedure evaluated every candidate. Of these, 993 (99.3\%) exceeded the target while passing the geometric checks. Among these 993 propeller-compatible candidates, the mean placement area was $0.7192\,\mathrm{m}^2$; the seven remaining candidates failed hull feasibility. All 27 fixed coordinates remained unchanged.

\begin{figure}[t]
\centering\includegraphics[width=2.75in]{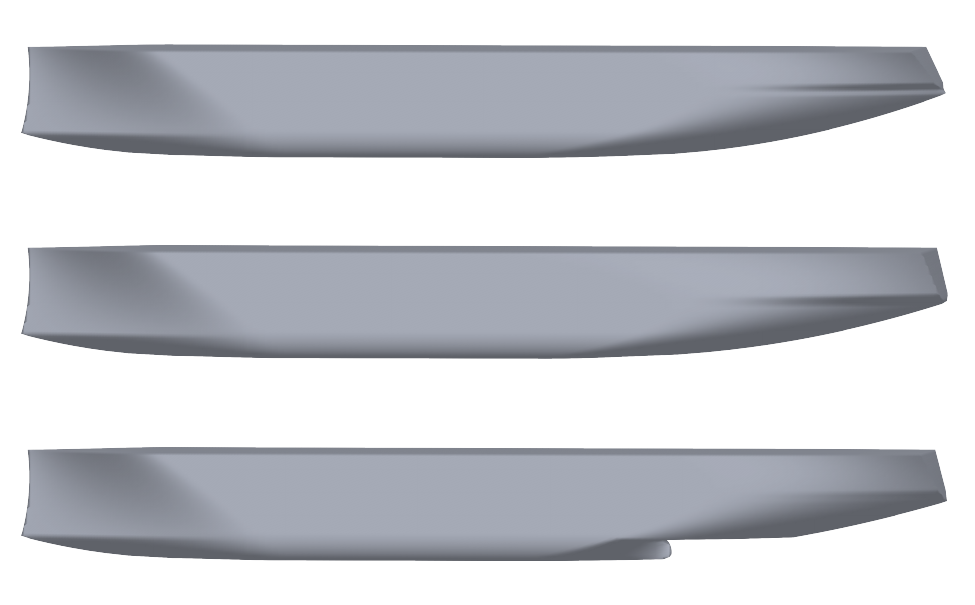}
\caption{Additional evaluated GenPFN stern candidates selected at the 10th, 50th and 90th percentiles of placement area among the 993 target-compatible generated designs. Only the permitted stern parameters differ from the initial hull.}
\label{fig_14}
\end{figure}

Figures~\ref{fig_12}--\ref{fig_14} show the maximum-area candidate ($1.079\,\mathrm{m}^2$) and the saved additional alternatives. Only the eighteen permitted stern controls vary from the row-17 parent. The examples demonstrate acquisition of propeller-compatible alternatives while preserving the bow and parallel midbody. They describe one parent, one context and one acquisition round, rather than an optimization study.

\section{Discussion}
Engineering exploration can focus on desired traits even when the corresponding region is difficult to delineate in the design variables. Labeled examples provide a practical way to describe this region through evaluated designs that possess or lack the traits of interest. GenPFN uses these examples to direct acquisition toward the desired class. This allows the acquisition goal to follow engineering requirements expressed through observed designs, without requiring an explicit mathematical description of the boundary separating desired and undesired designs.

The stern study illustrates this role when an existing design must accommodate an additional geometric requirement. Labeling examples by propeller compatibility directs exploration toward stern alternatives that provide space for placement while retaining the rest of the hull. These alternatives provide a basis for comparing design features associated with the desired trait and selecting designs for subsequent analysis. The engineering value lies in expanding the alternatives available for investigation within the region of interest; the case study itself evaluates geometric compatibility, while hydrodynamic and other vessel-performance assessments remain subsequent design tasks.

The regression study connects this exploration to learning about performance within the desired class. Because the validation designs belong to Class~1, improved prediction accuracy measures the value of acquired observations within the region that the engineering study seeks to understand. The acquired designs therefore serve two related purposes: they expand the available alternatives and supply evaluated responses for predictive modeling. Their usefulness depends on both their number and their distribution within the target region. The feasibility and regression comparisons consequently inform different acquisition decisions: obtaining more satisfactory alternatives and obtaining observations that improve performance prediction.

\subsection{Limitations}
A central limitation is the absence of systematic hyperparameter tuning. The step coefficient $\alpha$, clipping threshold $\lambda$, noise scale $\sigma$, probability threshold $P$, and number of updates determine how candidates move and spread through the design space. The study uses author-selected coefficients, and the sensitivity and long-run studies in Appendices~\ref{sec:step_sensitivity} and~\ref{app:long_noise} show that sampling behavior depends on the problem and trajectory length. Joint selection of these settings could better match the sampling process to the geometry of the desired region and the acquisition goal. Automatic or adaptive selection is an important direction for reducing manual choices and balancing class satisfaction, exploration, and computational effort.

The current method also guides generation through class membership alone. Although a class label can encode a performance requirement, the sampler does not directly use continuous objective values to distinguish better-performing designs within the desired class. Objective values are obtained through external evaluation and used for downstream regression training. A future extension could investigate guidance from PFN regression models to direct candidates toward desired performance levels or improved objective values, combined with classifier guidance for satisfying the target requirements. Such an extension would connect exploration of a desired class with performance-directed search within that class. Its development would require assessing the reliability of the regression guidance and how its strength affects the range of generated alternatives.

\section{Conclusion}
GenPFN expands small engineering datasets through class-targeted generation and external evaluation. A frozen TabPFN classifier guides candidates initialized between observed Class~1 designs, per-candidate L2 clipping controls the deterministic update, and Gaussian noise adds variation. Problem-specific constraints and objectives evaluate every proposal before feasible evaluated designs augment regression training.

Across the fourteen selected problems, GenPFN v1 has higher class satisfaction and lower realism distance than the tested CTGAN, TVAE and TabDDPM implementations. TVAE provides greater coverage and diversity, while CTGAN also provides greater dispersion. Relative to SMOTE, GenPFN has lower coverage distance and higher diversity; the adjusted class-satisfaction comparison is not significant. These results describe the balance between acquiring desired designs and spreading proposals through the design space.

Among observed paired regression fits, GenPFN-v1 augmentation reduces estimated AutoGluon $MSE$ by 62.1\% relative to context-only training. GenPFN has lower paired MSE than SMOTE, while TVAE and TabDDPM have lower paired MSE than GenPFN. Availability and the number of added feasible designs remain part of interpreting these workflow comparisons.

The configuration studies separate initialization, gating and guidance-free variation. At 100 updates, the pooled feasibility difference from interpolation plus noise is small and its pointwise interval spans zero. The supplementary long-run study shows larger guidance benefits for Two Circles and Compression Spring, with smaller separation on Hull and low absolute feasibility on Heat Exchanger. The local PFN activations and backbone comparison have strengths that depend on the endpoint and problem.

The stern application demonstrates targeted acquisition of geometric designs while preserving the rest of the parent hull. External evaluation verifies 99.3\% of the 1,000 candidates as target-compatible, compared with 1.4\% of the 2,000 random context designs.

GenPFN supports engineering exploration by using labeled examples to acquire additional evaluated designs, expanding the available alternatives and supporting more accurate performance prediction within the desired class.

\subsection{Future Work}
Future work will investigate systematic hyperparameter tuning and automatic or adaptive sampling-parameter selection to better balance class satisfaction, exploration, and computational effort. The current method uses class-based guidance alone. An additional direction is to investigate PFN regression models for guidance from continuous performance predictions. Combining this guidance with classifier guidance could support search for improved performance within the desired class. Other directions include multiclass and alternative PFNs and boundary- and uncertainty-focused acquisition.

\section*{Acknowledgment} %% ASME requests this exact spelling, singular.
The authors thank Rosen (Ting-Ying) Yu and Lyle Regenwetter for assistance in gathering the supplementary engineering test problems. The authors acknowledge the MIT Office of Research Computing and Data for providing high-performance computing resources that contributed to this work. Large language models assisted with language consolidation and grammar checks. The authors reviewed and edited the content and take full responsibility for the publication. The United States Department of Defense National Science and Engineering Graduate Fellowship funded this project. Code for GenPFN, the test problems, and the evaluation metrics is available at \url{https://github.com/noahbagz/GenPFN}.
%%%%%%%%%%%%%%%%%%%%%%%%%%%%%%%%%%%%%%%%%%%%%%%%%%%%%%%%%%%%%%%%%%%%%%

%%%%%%%%%%%%%%%  APPENDICES  %%%%%%%%%%%%%%%%%%%%%%%%%%%%%%%%%%%%%%%%%

%% Note that appendices will be "numbered" A, B, C, ... etc. Use \section, not \section*
%% Subsections need not be numbered, use \subsection*
%% The equation counter is reset for each appendix
%% Figures will be numbered consecutively
   %%% starting appendices
\newpage
\appendix
\noindent\textbf{Appendix Roadmap.} The appendices describe the additional engineering test problems, software environments, factorial configuration study, and guidance-free and controlled PFN comparisons. The final appendix reports the long-run noise-injection study.

\section{Additional Engineering Test Problems}
The benchmark contains Two Circles, Hull and twelve additional engineering problems. Saved Class~0 and Class~1 libraries supply the context subsets, while each fixed Class~1 reference set contains 20,000 evaluated designs. Ground-truth constraints assign the class labels; objectives and feasibility labels remain distinct from classifier predictions. Table~\ref{tab_6} summarizes all fourteen problems, including their parameter counts, context-library sizes and reference-set sizes.
The problems include the Ring example with the \emph{KNO1} objective~\citep{knowles2006parego} and established vehicle- and structural-design benchmarks~\citep{yu2024fast,jetton2024constraining,gu2001optimisation,gandomi2011mixed,golinski1973adaptive,yang2012bat,thanedar1995survey,koziel2011computational,sandgren1988nonlinear}.
\begin{table*}[t]
\caption{All fourteen engineering test problems, parameter counts, and Class~0 and Class~1 design-library sizes. The fixed Class~1 reference designs supply distance and regression validation.}
\label{tab_6}
\centering
\fitTableToWidth{\textwidth}{%
\begin{tabular}{@{}lllll@{}}
\toprule
Problem & Parameters & Class 0 library & Class 1 library & Reference designs \\
\midrule
Ring & 2 & 7,599 & 2,401 & 20,000 \\
Two Circles & 2 & 7,534 & 2,466 & 20,000 \\
JLH1 & 2 & 7,236 & 2,764 & 20,000 \\
JLH2 & 2 & 3,818 & 6,182 & 20,000 \\
Three-bar Truss & 2 & 7,820 & 2,180 & 20,000 \\
Pressure Vessel & 4 & 5,997 & 4,003 & 20,000 \\
Heat Exchanger & 8 & 998,712 & 1,288 & 20,000 \\
Concrete Beam & 3 & 8,406 & 1,594 & 20,000 \\
Welded Beam & 4 & 6,067 & 3,933 & 20,000 \\
Cantilever Beam & 10 & 994,180 & 5,820 & 20,000 \\
Compression Spring & 3 & 98,468 & 1,532 & 20,000 \\
Speed Reducer & 7 & 995,752 & 4,248 & 20,000 \\
Car Impact & 11 & 91,872 & 8,128 & 20,000 \\
Hull & 44 & 10,000 & 10,000 & 20,000 \\
\bottomrule
\end{tabular}
}
\end{table*}

\section{Software Environments and Implementation Settings}
\label{sec:software_environments}
TabPFN v1, TabPFN v3, and AutoGluon use separate software environments. Table~\ref{tab_7} reports the package versions and numerical precision recorded for these environments.
\begin{table*}[t]
\caption{Recorded software configurations for the TabPFN generation backends and AutoGluon regression environment.}
\label{tab_7}
\centering
\fitTableToWidth{\textwidth}{%
\begin{tabular}{@{}lll@{}}
\toprule
Environment & Package/setting & Recorded value \\
\midrule
TabPFN v1 backend & TabPFN & 0.1.10 \\
TabPFN v1 backend & PyTorch & 2.4.1+cu121 \\
TabPFN v1 backend & NumPy & 1.24.4 \\
TabPFN v1 backend & Precision & float32 \\
TabPFN v3 backend & TabPFN & 8.0.8 \\
TabPFN v3 backend & PyTorch & 2.5.1+cu121 \\
TabPFN v3 backend & NumPy & 2.5.0 \\
TabPFN v3 backend & Precision & float32 \\
AutoGluon regression record & Python & 3.10.16 \\
AutoGluon regression record & AutoGluon & 1.2 \\
AutoGluon regression record & NumPy & 1.26.4 \\
AutoGluon regression record & pandas & 2.2.3 \\
AutoGluon regression record & SciPy & 1.15.2 \\
\bottomrule
\end{tabular}
}
\end{table*}

TabPFN v1 uses two ensemble configurations, disabled external preprocessing, differentiable native normalization and input-gradient calculation. The labeled context remains fixed. Runs record model/proposal seeds, checkpoint hashes, and candidate ordering. The classifier remains frozen; no per-context neural-network optimization is performed. The v3 energy adapter is recorded separately from native probability guidance.
AutoGluon 1.2 uses a 15-second fit limit, training-only input and response scaling, feasibility-selected generated rows, fixed validation data, and eight CPUs per fit. No explicit fitting seed is supplied. Four workers execute independent fits. Missing fits, unsupported small contexts, and zero feasible yield are recorded explicitly.

\section{Factorial Configuration Study of the GenPFN Method}
\label{sec:ablation}
The factorial study uses TabPFN v1 and crosses direct and interpolated initialization with no, gated, and continuous probability guidance. Table~\ref{tab_8} defines the six configurations. Comparisons match the same problem, context size, class ratio, and subset identifier, retaining only contexts with results for all six configurations. A generated pool used in multiple comparisons is counted once rather than treated as an additional experiment.
\begin{table*}[t]
\caption{The $2\times3$ v1 configuration study. Initialization is direct from a Class~1 context design or interpolation between two such designs. None, gated and continuous guidance share the recorded sampling settings within each problem.}
\label{tab_8}
\centering
\fitTableToWidth{\textwidth}{%
\begin{tabular}{@{}lll@{}}
\toprule
Initialization & Guidance & Cell \\
\midrule
Interpolation & None & Interpolation + None \\
Interpolation & Gated & Interpolation + Gated \\
Interpolation & Continuous & Interpolation + Continuous \\
Direct & None & Direct + None \\
Direct & Gated & Direct + Gated \\
Direct & Continuous & Direct + Continuous \\
\bottomrule
\end{tabular}
}
\end{table*}

\subsection{Paired Factorial Results}
Table~\ref{tab_9} reports the six-cell factorial effects separately for eleven problems. The analysis uses $100f$ as the response, where $f$ is the feasibility fraction, and reports contrasts in percentage points. The guidance contrasts compare gated guidance with no guidance and continuous guidance with no guidance, averaged across direct and interpolated initialization. The interpolated--direct contrast averages over all three guidance settings.

Paired contrasts are averaged within context-size and class-ratio conditions and then across conditions. Approximate pointwise 95\% intervals use the condition-level variance calculation described in Experimental Methods. Estimates are reported separately for each problem; they are not pooled across the eleven problems or presented as multiplicity-adjusted tests.
\begin{table*}[t]
\caption{Six-cell v1 factorial results for eleven problems: feasibility differences in percentage points with approximate pointwise 95\% intervals from the condition-level variance calculation. Guidance contrasts average both initializations; interpolation--direct averages all guidance levels. Effects and intervals are reported separately for each problem. Bold point estimates mark the larger guidance-versus-none feasibility effect within each problem; the initialization contrast is reported separately.}
\label{tab_9}
\centering
\fitTableToWidth{\textwidth}{%
\begin{tabular}{@{}llll@{}}
\toprule
Problem & Gated--none & Continuous--none & Interpolation--direct \\
\midrule
Cantilever Beam & 0.001333 [-0.00128, 0.003947] & \textbf{1.18} [0.9585, 1.402] & 2.211 [1.944, 2.478] \\
Car Impact & 0.01161 [-0.001126, 0.02435] & \textbf{0.5566} [0.4121, 0.701] & 1.015 [0.8005, 1.23] \\
Concrete Beam & 0.004944 [0.003664, 0.006225] & \textbf{0.6095} [0.4562, 0.7628] & 0.6211 [0.5139, 0.7284] \\
JLH1 & 0.02778 [0.02116, 0.0344] & \textbf{0.3903} [0.3019, 0.4787] & 0.3578 [0.2927, 0.4229] \\
JLH2 & 0.01683 [0.01233, 0.02134] & \textbf{0.1251} [0.0946, 0.1555] & 0.1419 [0.1072, 0.1765] \\
Pressure Vessel & 0.05539 [-0.04511, 0.1559] & \textbf{0.4286} [0.2842, 0.5729] & 0.5546 [0.4477, 0.6615] \\
Ring & \textbf{3.183} [2.102, 4.264] & -12.72 [-15.17, -10.27] & -21.55 [-22.29, -20.81] \\
Speed Reducer & 0.0001667 [-6.574e-05, 0.0003991] & \textbf{2.453} [2.134, 2.772] & 1.924 [1.752, 2.096] \\
Three-bar Truss & 0.03761 [0.02851, 0.04672] & \textbf{0.4722} [0.3562, 0.5882] & 0.428 [0.3621, 0.494] \\
Two Circles & \textbf{6.965} [6.651, 7.28] & 3.318 [1.473, 5.163] & -9.09 [-9.521, -8.659] \\
Welded Beam & 0.004778 [0.002804, 0.006752] & \textbf{0.5845} [0.2931, 0.8759] & 0.6732 [0.5725, 0.7739] \\
\bottomrule
\end{tabular}
}
\end{table*}

The feasibility effects depend on the problem. Continuous guidance produces larger estimated gains than gated guidance for examples including Cantilever Beam and Car Impact, whereas gated guidance performs better on Ring and Two Circles. Interpolation also has different effects across problems, with a positive contrast for Cantilever Beam and negative contrasts for Ring and Two Circles. These are averaged factor effects, rather than comparisons of individual combined configurations.

\subsection{Representative Configuration Results}
Heat Exchanger, Hull, and Compression Spring contribute the interpolated guidance and control comparisons.

\begin{figure}[!th]
\centering\includegraphics[width=1.6in]{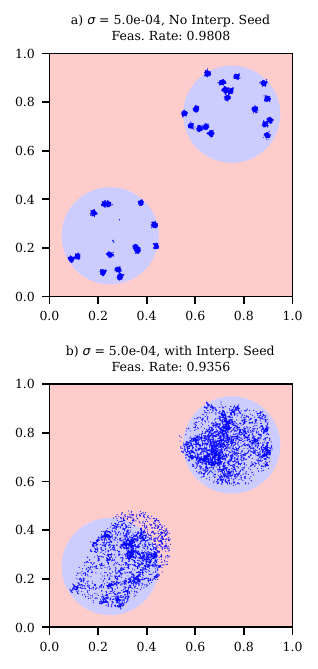}
\caption{Two Circles initialization using 64 context designs with equal class proportions, subset 0, and $\sigma=0.0005$. Direct starts above and interpolated starts below both use gated guidance.}
\label{fig_15}
\end{figure}

Figure~\ref{fig_15} compares direct and interpolated initialization for Two Circles at $\sigma=0.0005$.

\begin{figure*}[!th]
\centering\includegraphics[width=3.5in]{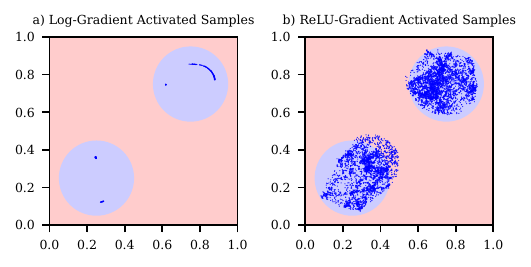}
\caption{Two Circles candidates generated using continuous and gated probability guidance from matching interpolated Class~1 starts. Both configurations use 64 context designs with equal class proportions and subset 0.}
\label{fig_16}
\end{figure*}

The continuous and gated candidate sets in Figure~\ref{fig_16} use the same 64-design context with equal class proportions and subset 0. Blue markers denote candidate proposals; feasibility comes from the saved external evaluation.

\begin{figure*}[!th]
\centering\includegraphics[width=5.5in]{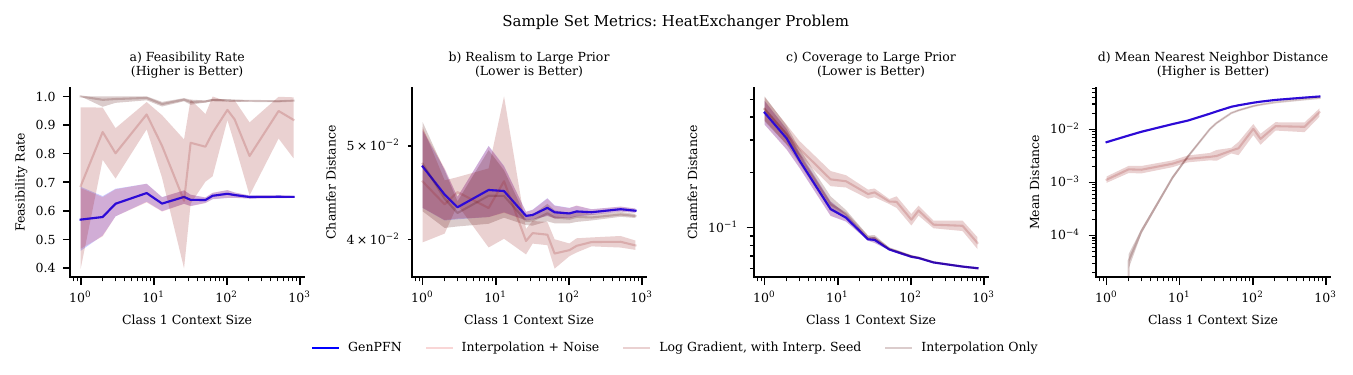}
\caption{Heat Exchanger sample-set metrics for interpolated v1 gated and continuous guidance, noise-only interpolation and pure interpolation. Panels retain feasibility, realism, coverage and nearest-neighbor dispersion roles.}
\label{fig_17}
\end{figure*}

\begin{figure}[!th]
\centering\includegraphics[width=2.5in]{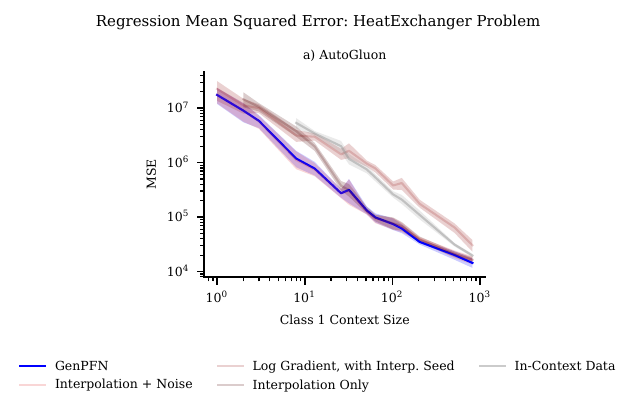}
\caption{Heat Exchanger AutoGluon validation mean squared error for the interpolated controls and probability-guidance arms, with context-only training as reference. Curves summarize observed regression fits.}
\label{fig_18}
\end{figure}

Figures~\ref{fig_17} and~\ref{fig_18} show Heat Exchanger sample-set metrics and AutoGluon validation MSE for interpolated guidance and control configurations.

Appendix~\ref{app:long_noise} separately examines cumulative guidance and noise from interpolated starts over longer trajectories.

\section{Guidance-Free and Controlled PFN Sensitivity Study}
\label{sec:pfn_sensitivity}
The controlled comparison uses the labeled context, matched starts and noise streams within each condition, and the common sampler. Interpolation alone and interpolation plus Gaussian noise provide shared controls. Table~\ref{tab_10} reports balanced contexts at total sizes 64 and 256 across subset IDs 0--9. Geometry metrics use every candidate in each pool; regression uses feasibility-selected designs and their evaluated responses.
\begin{table*}[t]
\caption{Controlled v1 geometry at balanced total context sizes 64 and 256, averaged over ten subsets. Feasibility is a fraction of all 5,000 proposals. Distances use every proposal in normalized coordinates. Continuous probability guidance and energy activation use the labeled context within the common sampler. Bold denotes the best displayed value per metric within each problem and context size, including ties.}
\label{tab_10}
\centering
\fitTableToWidth{\textwidth}{%
\begin{tabular}{@{}lllllll@{}}
\toprule
Problem & Context & Method & Feasibility ($\uparrow$) & Realism ($\downarrow$) & Coverage ($\downarrow$) & Diversity ($\uparrow$) \\
\midrule
Hull & 64 & GenPFN & 0.9049 & \textbf{1.447} & \textbf{1.741} & \textbf{0.1182} \\
Hull & 64 & Interpolation & \textbf{0.906} & \textbf{1.447} & \textbf{1.741} & 0.1169 \\
Hull & 64 & Interpolation + noise & 0.9049 & \textbf{1.447} & \textbf{1.741} & \textbf{0.1182} \\
Hull & 64 & Local continuous guidance & 0.9025 & \textbf{1.447} & \textbf{1.741} & 0.1181 \\
Hull & 64 & Local energy activation & 0.9029 & \textbf{1.447} & \textbf{1.741} & 0.118 \\
Hull & 256 & GenPFN & 0.8956 & 1.442 & 1.643 & \textbf{0.4399} \\
Hull & 256 & Interpolation & 0.8969 & 1.442 & 1.643 & 0.4396 \\
Hull & 256 & Interpolation + noise & 0.8955 & 1.442 & 1.643 & \textbf{0.4399} \\
Hull & 256 & Local continuous guidance & \textbf{0.897} & \textbf{1.441} & \textbf{1.642} & 0.4389 \\
Hull & 256 & Local energy activation & 0.8966 & \textbf{1.441} & 1.643 & 0.4392 \\
Two Circles & 64 & GenPFN & 0.9115 & 0.005206 & \textbf{0.0129} & 0.002956 \\
Two Circles & 64 & Interpolation & 0.7598 & 0.02043 & 0.01449 & 0.003156 \\
Two Circles & 64 & Interpolation + noise & 0.7591 & 0.02044 & 0.01311 & \textbf{0.003495} \\
Two Circles & 64 & Local continuous guidance & \textbf{1} & \textbf{0.001689} & 0.0957 & 0.000418 \\
Two Circles & 64 & Local energy activation & 0.4192 & 0.1819 & 0.09827 & 0.0006075 \\
Two Circles & 256 & GenPFN & 0.9747 & 0.002065 & \textbf{0.006095} & 0.003196 \\
Two Circles & 256 & Interpolation & 0.7482 & 0.0212 & 0.006537 & 0.003917 \\
Two Circles & 256 & Interpolation + noise & 0.7471 & 0.02121 & 0.006266 & \textbf{0.004005} \\
Two Circles & 256 & Local continuous guidance & \textbf{1} & \textbf{0.001772} & 0.02193 & 0.002279 \\
Two Circles & 256 & Local energy activation & 0.07576 & 0.1869 & 0.1233 & 0.001451 \\
Heat Exchanger & 64 & GenPFN & 0.6384 & 0.04243 & \textbf{0.08571} & \textbf{0.02185} \\
Heat Exchanger & 64 & Interpolation & \textbf{0.9805} & 0.04186 & 0.08775 & 0.01323 \\
Heat Exchanger & 64 & Interpolation + noise & 0.6384 & 0.04243 & \textbf{0.08571} & \textbf{0.02185} \\
Heat Exchanger & 64 & Local continuous guidance & 0.8379 & 0.04063 & 0.1555 & 0.003159 \\
Heat Exchanger & 64 & Local energy activation & 0.8529 & \textbf{0.04043} & 0.1571 & 0.002652 \\
Heat Exchanger & 256 & GenPFN & 0.6554 & 0.04278 & \textbf{0.06829} & \textbf{0.03337} \\
Heat Exchanger & 256 & Interpolation & \textbf{0.985} & 0.04224 & 0.06867 & 0.02935 \\
Heat Exchanger & 256 & Interpolation + noise & 0.6554 & 0.04278 & \textbf{0.06829} & \textbf{0.03337} \\
Heat Exchanger & 256 & Local continuous guidance & 0.9197 & 0.03942 & 0.1239 & 0.006625 \\
Heat Exchanger & 256 & Local energy activation & 0.8687 & \textbf{0.03916} & 0.1296 & 0.004457 \\
\bottomrule
\end{tabular}
}
\end{table*}

In Two Circles, continuous guidance achieves higher feasibility, while GenPFN provides lower coverage distance and greater diversity. Hull shows little separation among the configurations. For Heat Exchanger, pure interpolation retains higher feasibility than the updated configurations. These comparisons describe the balanced context sizes displayed in Table~\ref{tab_10}.

\subsection{Controlled TabPFGen and TabEBM Energy Variants}
The local continuous probability guidance, labeled TabPFGen in the exports, uses the same frozen conditional Class~1 probability as GenPFN without the gate. Equation~\ref{eq_15} defines its energy for a query batch:
\begin{align}
E_{\mathrm{continuous}}&=-\sum_{i=1}^{B}\log\!\left[p_\theta(\boldsymbol{z}_i\mid\mathcal{D}_{\mathrm{ic}})+\delta\right].
\label{eq_15}
\end{align}
The local TabEBM activation uses the negative log-partition of the recorded class scores on the same labeled context, as defined in Equation~\ref{eq_16}:
\begin{equation}
E_{\mathrm{energy}}=-\sum_{i=1}^{B}\log\!\left[\sum_{k=0}^{1}\exp f_k(\boldsymbol{z}_i\mid\mathcal{D}_{\mathrm{ic}})\right].
\label{eq_16}
\end{equation}
The v3 comparison uses the mean-log-partition score adapter. These labels identify local activation comparisons~\citep{ma2023tabpfgen,margeloiu2024tabebm}; no synthetic negative rows, reverse-context term or class-specific surrogate context is introduced. All activations use the same per-candidate L2 clipping, fixed horizon and problem-specific coefficients.
Table~\ref{tab_11} reports observed successful regression fits for the representative conditions. These means describe augmentation with method-specific feasible yields. The lowest displayed MSE comes from context-only training for Hull and GenPFN for Two Circles at both context sizes. For Heat Exchanger, interpolation plus noise gives the lowest displayed MSE at context size 64, while GenPFN gives the lowest at 256.
\begin{table*}[t]
\caption{Mean AutoGluon validation $MSE$ for observed fits under the balanced-context conditions in Table~\ref{tab_10}. Each method augments the context with its feasibility-selected generated designs, so added row counts can differ. Missing fits are not imputed. Fit counts are provided in the supporting data. Bold denotes the lowest displayed mean MSE within each problem and context size, including ties.}
\label{tab_11}
\centering
\fitTableToWidth{\textwidth}{%
\begin{tabular}{@{}llrr@{}}
\toprule
Problem & Method & Context 64 & Context 256 \\
\midrule
Hull & Context only & \textbf{0.03612} & \textbf{0.01986} \\
Hull & GenPFN & 0.08273 & 0.1269 \\
Hull & Interpolation & 0.3061 & 0.0943 \\
Hull & Interpolation + noise & 0.1315 & 0.0222 \\
Hull & Local continuous guidance & 0.1243 & 0.06464 \\
Hull & Local energy activation & 0.1205 & 0.1253 \\
Two Circles & Context only & 43.48 & 18 \\
Two Circles & GenPFN & \textbf{12.1} & \textbf{8} \\
Two Circles & Interpolation & 12.7 & 8.797 \\
Two Circles & Interpolation + noise & 12.72 & 8.211 \\
Two Circles & Local continuous guidance & 42.7 & 11.25 \\
Two Circles & Local energy activation & 35.44 & 17.82 \\
Heat Exchanger & Context only & 1.171e+06 & 2.095e+05 \\
Heat Exchanger & GenPFN & 3.157e+05 & \textbf{6.187e+04} \\
Heat Exchanger & Interpolation & 3.132e+05 & 6.887e+04 \\
Heat Exchanger & Interpolation + noise & \textbf{3.05e+05} & 6.543e+04 \\
Heat Exchanger & Local continuous guidance & 1.65e+06 & 4.223e+05 \\
Heat Exchanger & Local energy activation & 1.691e+06 & 5.339e+05 \\
\bottomrule
\end{tabular}
}
\end{table*}

\section{Sensitivity to the Step Transformation}
\label{sec:step_sensitivity}
Preliminary testing compared two combinations of the step coefficient $\alpha$ and clipping threshold $\lambda$ (Table~\ref{tab_13}). For Two Circles, $\alpha=0.1$ and $\lambda=2$ produced higher feasibility and lower coverage distance than $\alpha=0.001$ and $\lambda=0.1$. Hull and Heat Exchanger showed little change between settings. These results illustrate the potential benefit of problem-specific hyperparameter selection.
\subsection{Symmetric Step Transformation}
Equation~\ref{eq_17} defines the deterministic map for a raw descent gradient $\boldsymbol{g}$:
\begin{equation}\label{eq_17}
\mathcal{S}_{\lambda}(\boldsymbol{g})=\alpha\lambda\frac{\boldsymbol{g}}{\max(\lambda,\lVert\boldsymbol{g}\rVert_2)}.
\end{equation}
Reversing the gradient reverses the deterministic step without changing its magnitude, and a zero gradient produces a zero deterministic step. The step norm is bounded by $\alpha\lambda$; Gaussian perturbations are added separately and are not subject to that bound.
\begin{table*}[t]
\caption{Preliminary step-strength comparison using per-candidate L2 clipping. Entries are mean feasibility fractions, coverage distances, and deterministic step norms for the listed coefficients. The problem-specific benchmark settings are given in Experimental Methods. Bold denotes the best displayed feasibility and coverage within each problem, including ties.}
\label{tab_13}
\centering
\fitTableToWidth{\textwidth}{%
\begin{tabular}{@{}llllll@{}}
\toprule
Problem & $\alpha$ & $\lambda$ & Feasibility ($\uparrow$) & Coverage ($\downarrow$) & Mean step norm \\
\midrule
Heat Exchanger & 0.001 & 0.1 & \textbf{0.6583} & \textbf{0.06914} & 0 \\
Heat Exchanger & 0.1 & 2 & \textbf{0.6583} & \textbf{0.06914} & 0 \\
Hull & 0.001 & 0.1 & \textbf{0.9024} & \textbf{1.644} & 5.173e-08 \\
Hull & 0.1 & 2 & 0.902 & \textbf{1.644} & 1.2e-06 \\
Two Circles & 0.001 & 0.1 & 0.7409 & 0.006497 & 2.346e-05 \\
Two Circles & 0.1 & 2 & \textbf{0.9805} & \textbf{0.006272} & 0.0006416 \\
\bottomrule
\end{tabular}
}
\end{table*}

\section{Long-Run Noise-Injection Study}
\label{app:long_noise}
The long-run study examines how cumulative guidance and noise affect feasibility beyond the 100-update benchmark. It follows gated v1 GenPFN and a noise-only reference to 30,000 updates using ten balanced 256-design contexts per problem and 5,000 candidates per arm/context. The recorded checkpoints are 0, 1, 3, 10, 30, 100, 300, 1,000, 3,000, 10,000 and 30,000 updates. Checkpoint zero contains the interpolated candidates before the initial Gaussian perturbation. Subsequent checkpoints include that perturbation and the stated number of updates. The primary endpoint is fixed at 30,000 updates.

The two arms share contexts, initial candidates and the Gaussian noise law, but use different subsequent noise draws. The noise-only arm advances between checkpoints by adding independent Gaussian displacements representing the accumulated noise over each interval. Comparisons are paired by context; candidate rows and successive checkpoints are not independent replications. Means summarize the ten contexts, and pointwise confidence intervals use paired-context bootstrap resampling. The intervals are not multiplicity-adjusted.

\begin{figure*}[!th]
\centering\includegraphics[width=6.5in]{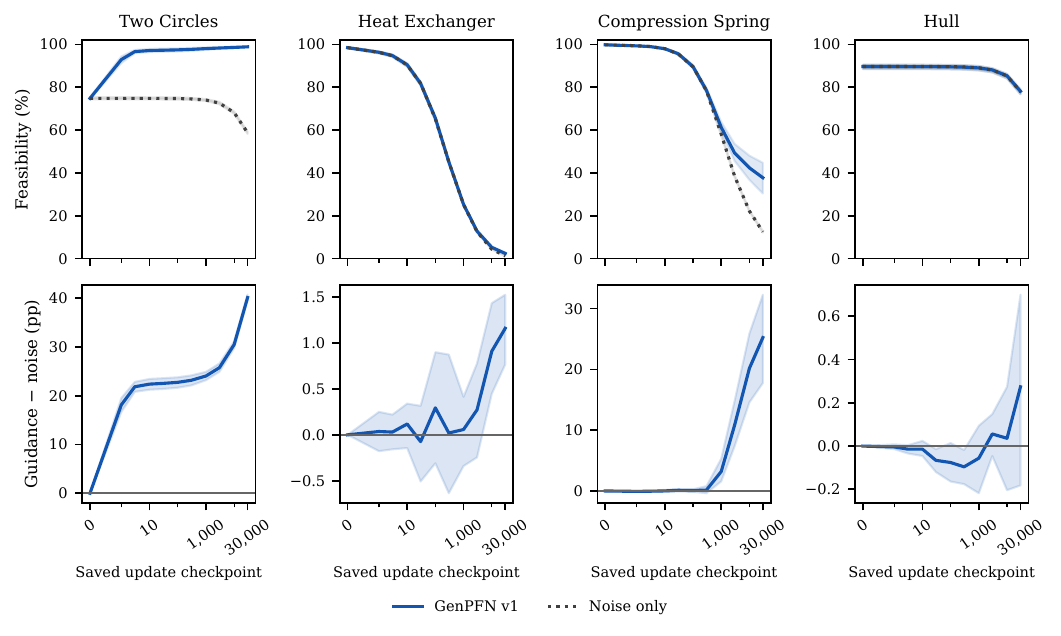}
\caption{Gated GenPFN and noise-only feasibility trajectories using the stated sampling settings. Bands are pointwise paired-context bootstrap intervals; the 30,000-update endpoint was specified in advance.}
\label{fig_19}
\end{figure*}

Table~\ref{tab:long_endpoint} reports feasibility for both arms and their paired difference at the 30,000-update endpoint. Figure~\ref{fig_19} shows the trajectories across checkpoints. At 30,000 updates, guidance produces larger feasibility gains for Two Circles and Compression Spring. Heat Exchanger retains low absolute feasibility in both arms, while Hull's interval spans zero. Future work will examine problem-specific hyperparameter tuning to strengthen GenPFN's feasibility benefit over noise-only sampling.
\begin{table*}[t]
\caption{Primary 30,000-update feasibility endpoint over ten balanced 256-row contexts and 5,000 candidates per arm/context. Difference is guided minus noise-only in percentage points, with the pointwise 95\% paired-context bootstrap interval. Starts and noise laws match; subsequent random draws differ. Intervals are not adjusted across checkpoints or problems. Bold denotes the higher displayed mean feasibility within each problem.}
\label{tab:long_endpoint}
\centering
\fitTableToWidth{\textwidth}{%
\begin{tabular}{@{}llll@{}}
\toprule
Problem & GenPFN v1 (\%) & Noise-only (\%) & Difference (pp) [95\% CI] \\
\midrule
Two Circles & \textbf{98.89} & 58.77 & 40.12 [39.53, 40.70] \\
Heat Exchanger & \textbf{2.384} & 1.224 & 1.160 [0.766, 1.526] \\
Compression Spring & \textbf{37.74} & 12.50 & 25.23 [17.75, 32.28] \\
Hull & \textbf{78.14} & 77.86 & 0.274 [-0.182, 0.700] \\
\bottomrule
\end{tabular}
}
\end{table*}

\FloatBarrier
\bibliographystyle{asmejour}   %% .bst file that follows ASME journal format. Do not change.

\bibliography{biblio} %% <=== change this to name of your bib file

%%%%%%%%%%%%%%%%%%%%%%%%%%%%%%%%%%%%%%%%%%%%%%%%%%%%%%%%%%%%%%%%%%%%%%

%% To omit final list of figures and tables, use the class option [nolists]

\end{document}